\documentclass[journal]{IEEEtran}
\usepackage{amsmath,amsfonts}
\usepackage{algorithmic}
\usepackage{algorithm}
\usepackage{array}
\usepackage[caption=false,font=normalsize,labelfont=sf,textfont=sf]{subfig}
\usepackage{textcomp}
\usepackage{stfloats}
\usepackage{url}
\usepackage{verbatim}
\usepackage{graphicx}
\usepackage{cite}
\usepackage{color,xcolor}
\usepackage{amssymb}
\usepackage{bm}
\usepackage{multirow}
\usepackage{hyperref}
\hypersetup{
	colorlinks=true,
	linkcolor=red}

\usepackage[capitalize]{cleveref}

\begingroup
\renewcommand{\arraystretch}{1} 

\usepackage{booktabs}
\begin{document}

\title{Mixed-granularity Implicit Representation for Continuous Hyperspectral Compressive Reconstruction}

\author{Jianan Li$^{*}$, Huan Chen$^{*}$, Wangcai Zhao$^{\dagger}$, Rui Chen, Tingfa Xu$^{\dagger}$ }

\maketitle

\begin{abstract}

Hyperspectral Images (HSIs) are crucial across numerous fields but are hindered by the long acquisition times associated with traditional spectrometers. The Coded Aperture Snapshot Spectral Imaging (CASSI) system mitigates this issue through a compression technique that accelerates the acquisition process. However, reconstructing HSIs from compressed data presents challenges due to fixed spatial and spectral resolution constraints. This study introduces a novel method using implicit neural representation for continuous hyperspectral image reconstruction. We propose the Mixed Granularity Implicit Representation (MGIR) framework, which includes a Hierarchical Spectral-Spatial Implicit Encoder for efficient multi-scale implicit feature extraction. This is complemented by a Mixed-Granularity Local Feature Aggregator that adaptively integrates local features across scales, combined with a decoder that merges coordinate information for precise reconstruction. By leveraging implicit neural representations, the MGIR framework enables reconstruction at any desired spatial-spectral resolution, significantly enhancing the flexibility and adaptability of the CASSI system. Extensive experimental evaluations confirm that our model produces reconstructed images at arbitrary resolutions and matches state-of-the-art methods across varying spectral-spatial compression ratios. The code will be released at~\url{https://github.com/chh11/MGIR}.
\end{abstract}


\section{Introduction}

\IEEEPARstart{H}{yperspectral} Images (HSIs) are widely used across various fields~\cite{jolly2022hyperspectral, 10313066, 10475351}. Traditional imaging spectrometers, however, require extended periods to capture multiple spectral bands~\cite{zhang2021deeply}. To overcome this limitation, Coded Aperture Snapshot Spectral Imaging (CASSI) technology has been developed~\cite{tang2022single, 10261266}. This approach uses a novel compression strategy along the spectral dimension to significantly reduce imaging time, thereby increasing efficiency.

\begin{figure}[tbp]
\centering
\includegraphics[width=\linewidth]{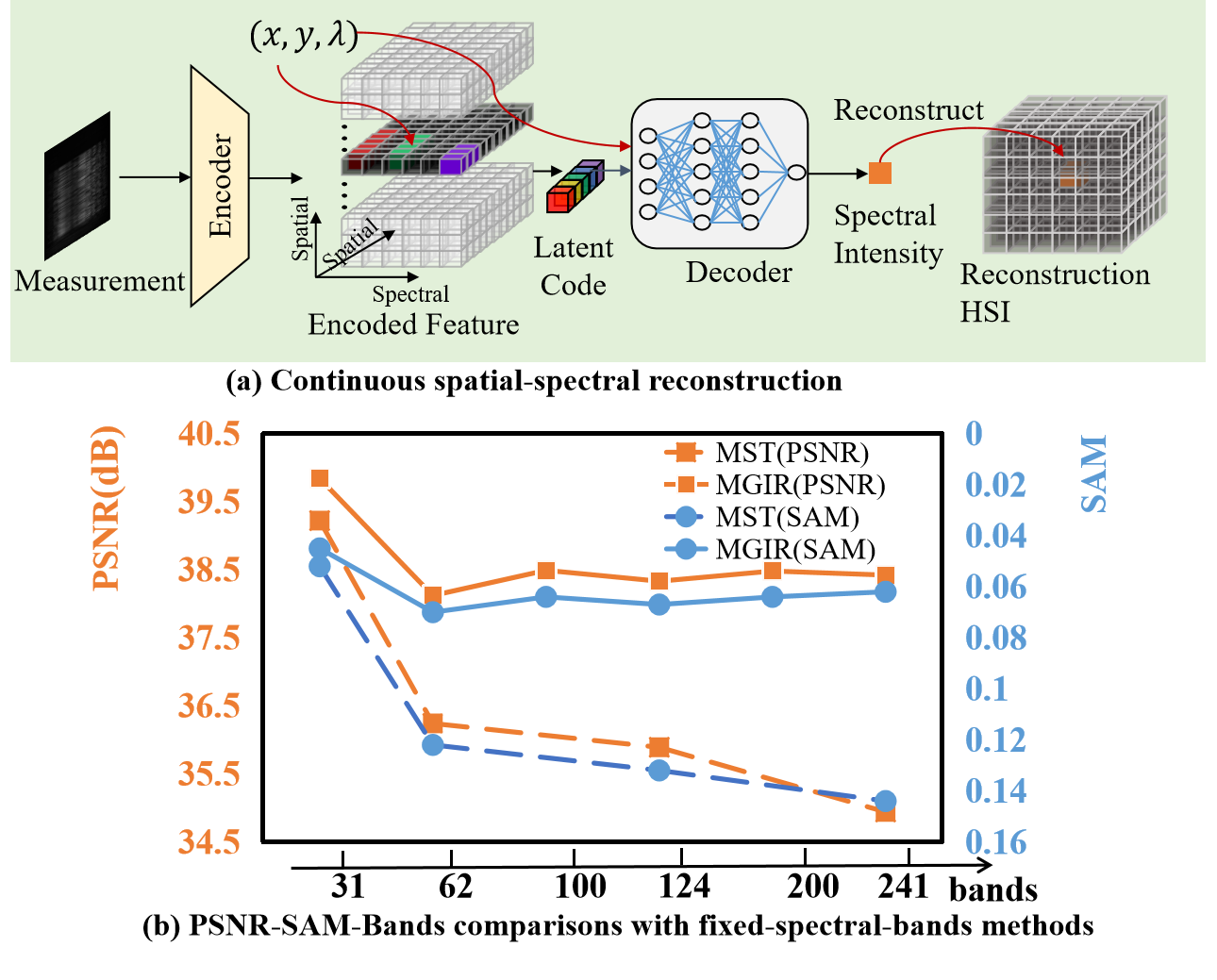}
\caption{(a) Principle of hyperspectral image Implicit Neural Representation involves encoding compressed measurements into a feature space. Given any spatial-spectral coordinate and corresponding features extracted by the encoder from the input, it decodes the spectral intensity at that coordinate. (b) Our Mixed-Granularity Implicit Representation demonstrates enhanced reconstruction quality across continuous spectral bands, outperforming the state-of-the-art fixed-band method MST~\cite{cai2022mask}.}

\label{fig1}
\end{figure}

The CASSI system depends on reconstruction algorithms to accurately convert two-dimensional (2D) compressed images into three-dimensional (3D) hyperspectral data cubes. Recently, deep learning models have emerged as a promising, automated method for achieving high-quality hyperspectral image reconstruction~\cite{wang2021tsa, meng2020gap, cai2022degradation}.
However, existing reconstruction methods are limited by the detector size of the CASSI system in terms of spatial resolution and typically can only reconstruct a fixed number of discrete spectral bands in the spectral dimension. These limitations affect the flexibility and applicability of the models in handling different compression ratios and various imaging conditions.

To address existing limitations, we integrate Implicit Neural Representation (INR) into hyperspectral image reconstruction for continuous output~\cite{zhang2021holistic, du2021neural, wang2021neus, chen2021learning}. INR employs a deep neural network to map object coordinates into a continuous spectral-spatial signal.
In practice, once the CASSI system compresses the continuous hyperspectral signal, it is stored as two-dimensional measurements. Our method feeds these measurements and spatial-spectral coordinates into the INR model, predicting signal intensity at each coordinate point. Significantly, by handling continuous coordinates and signals, our approach enables the reconstructed hyperspectral image to be displayed at any spatial and spectral resolution, enhancing the algorithm's flexibility.

However, implementing continuous reconstruction presents significant challenges. Hyperspectral image reconstruction typically relies on 2D backbone networks that inadequately capture joint spatial-spectral features, making them suboptimal for implicit representation. Although 3D backbone networks enhance performance by increasing parameters, this leads to substantial computational overhead and slower training and inference processes. Consequently, there is an urgent need for a lightweight 3D backbone network that maintains high performance while reducing computational costs. Moreover, HSIs display diverse scale characteristics across different spatial-spectral coordinates. Traditional 3D backbone networks, with their single-scale architectures, fail to adequately capture these variations, limiting reconstruction quality. Effective extraction and integration of multi-scale features are therefore crucial for accurate hyperspectral image reconstruction.

To address the challenges described, we introduce the Mixed-Granularity Implicit Representation (MGIR), a new approach for continuous hyperspectral compressed reconstruction. The MGIR framework consists of three key components: the Hierarchical Spectral-Spatial Implicit Encoder (HSSIE) for efficient feature extraction, the Mixed-Granularity Local Feature Aggregator (MGLFA) that adaptively integrates multi-scale features, and a decoder that uses parameterized implicit neural representation to reconstruct hyperspectral image.

First, we propose a Hierarchical Spectral-Spatial Implicit Encoder (HSSIE) for efficient spectral-spatial feature extraction. The fundamental unit of HSSIE is the Spectral-Spatial Depthwise Convolution module (SSDW), characterized by its minimal parameters. The SSDW employs two depthwise convolution layers to independently extract spatial and spectral channel features. By stacking patch embedding and SSDW modules to form a hierarchical structure, HSSIE captures a series of multi-granularity features from the input image.

After extracting features from the entire input using an encoder, it is essential to obtain multi-granularity features at a specific target location to reconstruct the spatial and spectral information at that position. To adaptively fuse features of varying granularity based on the characteristics of the target location, we introduce a Mixed-Granularity Local Feature Aggregator. This aggregator collects local features of multiple granularities around the target location using a divide-and-conquer approach. Specifically, it applies multi-head attention, where attention heads are divided into multiple groups, each responsible for attending to information within a specific range. The outputs of all groups are then merged to obtain the final mixed-granularity features.

As a result, the Hierarchical Spectral-Spatial Implicit Encoder and the Mixed-Granularity Local Feature Aggregator collaboratively map compressed measurements from a specific target location into a multi-granularity implicit local representation. Subsequently, a decoder is utilized, which processes this representation in conjunction with the coordinates of the target location. This decoder parameterizes the implicit representation, enabling the prediction of the reconstructed spatial-spectral signal at that point.

Our approach incorporates continuous learning for HSI reconstruction, effectively overcoming the inherent spatial-spectral resolution limitations of traditional methods. By enabling reconstruction at arbitrary and extreme compression ratios beyond the scope of the training samples, it facilitates reconstruction at any desired spatial-spectral resolution. This capability significantly enhances the adaptability and application potential of the CASSI system across diverse scenarios.

Extensive experiments across multiple datasets clearly demonstrate the effectiveness of our proposed method, both qualitatively and quantitatively. To sum up, this work makes the following contributions: follows:
\begin{itemize}

\item The proposed Mixed-Granularity Implicit Representation marks the first attempt at continuous HSI reconstruction using implicit representation techniques, achieving reconstruction at any desired spatial-spectral resolution.
\item We have developed a lightweight Hierarchical Spectral-Spatial Implicit Encoder that efficiently extracts spatial-spectral features with high computational efficiency.
\item We introduce a Multi-Scale Local Feature Aggregator that adaptively integrates features of varying granularities, thereby enhancing the model's capacity to process intricate image details.
\end{itemize}

\section{Related Work}

\noindent\textbf{Hyperspectral Image Reconstruction.}
Deep learning-based models have been widely employed in hyperspectral compressive reconstruction, demonstrating significant advancements. GAP-Net~\cite{meng2020gap} utilizes deep unfolding to reconstruct snapshot compressive images by adapting wide projection simulation signals to accommodate various systems. $\lambda$-Net~\cite{miao2019net} and TSA-Net~\cite{wang2021tsa} are two prominent end-to-end frameworks that leverage deep CNNs to learn effective signal mapping functions. RDFNet~\cite{10017363} further advances spectral SCI by dynamically adjusting regional features and learning adaptive thresholds, enabling more precise reconstructions. Similarly, the DAUHST~\cite{cai2022degradation} algorithm addresses data mismatches by dividing the task into data and prior modules within a deep unfolding framework.  
However, these existing learning-based approaches primarily target reconstruction of discrete spectral bands, failing to capture the inherent continuity of the 3D hyperspectral signal across spatial and spectral dimensions. In contrast, our proposed method introduces a multi-granularity latent feature learning framework specifically designed for continuous reconstruction. By leveraging the intrinsic correlations in hyperspectral data, our approach achieves seamless reconstruction across all spatial and spectral resolutions, setting a new standard that addresses the limitations of prior methods.

\noindent\textbf{Implicit Neural Representation.}
Implicit Neural Representation characterizes objects by correlating universal functions with neural networks. This methodology has been increasingly applied to fields such as 3D scene modeling~\cite{sitzmann2019scene, jiang2020local}. Mildenhall~\textit{et al.}~\cite{mildenhall2021nerf} illustrated how to capture the implicit representation of a single scene from multi-angle photographs to generate new perspectives. Branching into images, LIIF~\cite{chen2021learning} introduces the concept of continuous spatial scaling. However, the spectral dimension, another inherent continuous signal, frequently remains overlooked. NeSR realizes spectral continuity, facilitating the transition from RGB to hyperspectral imaging, although it is constrained by the necessity for specific parameters for spectral bands.

\noindent\textbf{Multi-head Attention Mechnism.}
Transformers, celebrated for their ability to handle long-range spatial interactions, have seen widespread use across various disciplines~\cite{han2021transformer}. Notably, MST~\cite{cai2022mask} stands as the inaugural transformer-based model dedicated to hyperspectral image reconstruction, where it treats spectral graphs as tokens and employs self-attention across the spectral dimension. Furthermore, Wang \textit{et al.}\cite{wang2022uformer} introduced UFormer, which utilizes the Swin Transformer's\cite{liu2021swin} building blocks for natural image restoration. However, existing transformer models often densely sample tokens-many of which may carry minimal information-and perform multi-head self-attention among these tokens, which limits the size of the model's receptive field. Typically, these methods are constrained by a fixed and small receptive field, which restricts the model's capacity to adaptively and flexibly cover the entire feature map. Ren \textit{et al.}~\cite{ren2021shunted} enhanced image feature acquisition by integrating spectral-spatial dual-branch strategies with patches of varying sizes. Drawing inspiration from CrossViT~\cite{chen2021crossvit}, we propose a scale-aware local attention mechanism that enables multi-granularity feature fusion, facilitating continuous hyperspectral reconstruction.

\section{Method}
\subsection{Problem Revisit}
\noindent\textbf{CASSI System.} 
The CASSI system captures 3D hyperspectral imagery in a single snapshot. Conventional reconstruction approaches, which perceive hyperspectral images as collections of discrete spatial pixels and spectral bands, often produce segmented reconstructions that overlook the inherent continuity of natural signals. In contrast, we treat the 3D hyperspectral images captured by CASSI as continuous functions across spectral-spatial coordinates, guiding our method for continuous hyperspectral image reconstruction.

\begin{figure}[tbp]
\centering
\includegraphics[width=\linewidth]{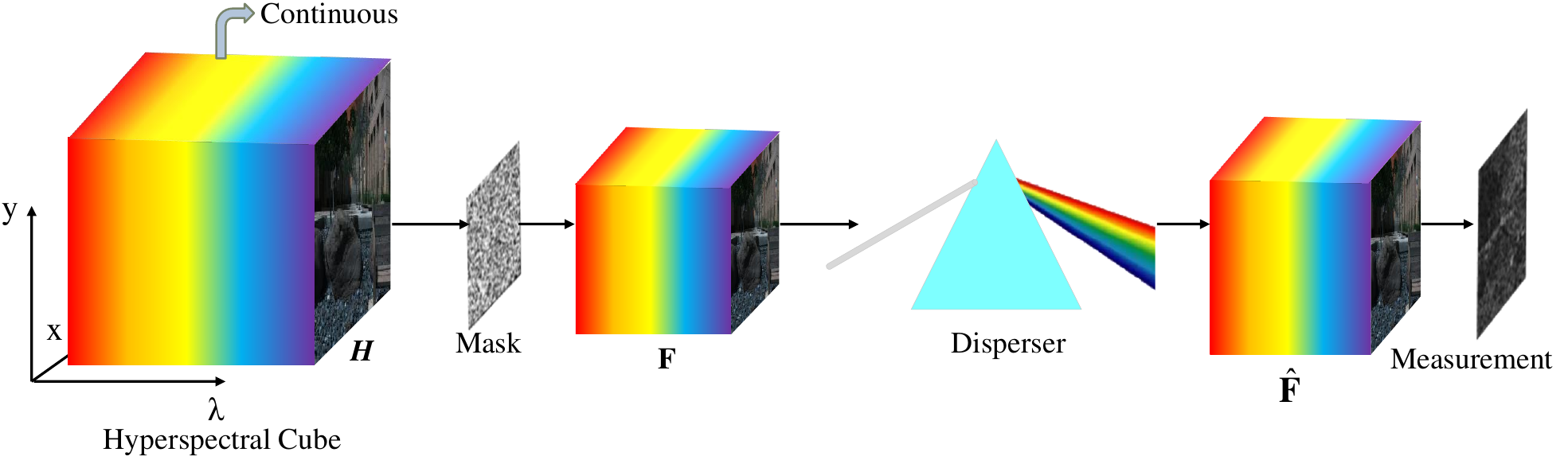}
\caption{Schematic of the CASSI System: Continuous spectral cubes are encoded by a mask, sorted by a disperser, and subsequently compressed into a 2D measurement.}
\label{fig2}
\end{figure}

\cref{fig2} illustrates the CASSI system, which captures a continuous real visual scene denoted as $\boldsymbol L \in {\mathbb{R}^{\rm H \times \rm W \times \rm{ \lambda}}} $, where $\rm H$, $\rm W$, and $ \lambda$ represent the height, width, and number of wavelengths of the hyperspectral image, respectively.
The scene passes through an objective lens and is spatially encoded by a fixed mask ${\boldsymbol M} \in {\mathbb{R}^{\rm H \times \rm W}}$, expressed as:
\begin{equation}
  \boldsymbol{F} = {\boldsymbol L_{\downarrow }  \odot {\boldsymbol M}},
  \label{eq:1}
\end{equation}
where $\downarrow$ is spatial discrete downsampling and $\odot$ signifies element-wise multiplication. 

Subsequently, a disperser spreads $\boldsymbol{F}$ along the y-axis to different spatial locations based on their wavelengths, resulting in the tilted cube $\boldsymbol{F}^{'} \in {\mathbb{R}^{\rm H \times \rm {W^{'}} \times \rm { \lambda }}}$, where $\rm {W^{'}}$ indicates the expanded width after dispersion along the y-axis. This transformation is described by:
\begin{equation}
  {\boldsymbol{F}^{'}}(u,v, \lambda ) = \boldsymbol{F}(x,y + \int_{\lambda_{\rm{min}}}^{\lambda_{\rm{max}}} {f_{\lambda }}d\lambda ,\rm{\lambda}),
  \label{eq:2}
\end{equation}
where $(u,v)$ signifies the coordinates on the detector plane, and the integral $\int_{\lambda_{\rm{min}}}^{\lambda_{\rm{max}}} {f_{\lambda}}d{\lambda }$ represents the spatial across a continuous range of wavelengths from ${\lambda_{\rm{min}}}$ to ${\lambda_{\rm{max}}}$.

Finally, the compressed measurement at the detector, denoted as $g(u, v)$, is modeled by the equation:
\begin{equation}
  g(u,v) = \int_{\lambda_{\rm{min}}}^{\lambda_{\rm{max}}} {\boldsymbol{F}^{'}} (u,v, \lambda )d{\lambda}.
  \label{eq:3}
\end{equation}
The sensor integrates all the light in the wavelength range $[\lambda_{\rm{min}},\lambda_{\rm{max}}]$. Thus, the 2D measurement input for CASSI reconstruction encompasses both spatial and spectral continuous signals, forming the core hypothesis of our study.

\begin{figure*}[t]
  \centering
    \includegraphics[width=0.95\linewidth]{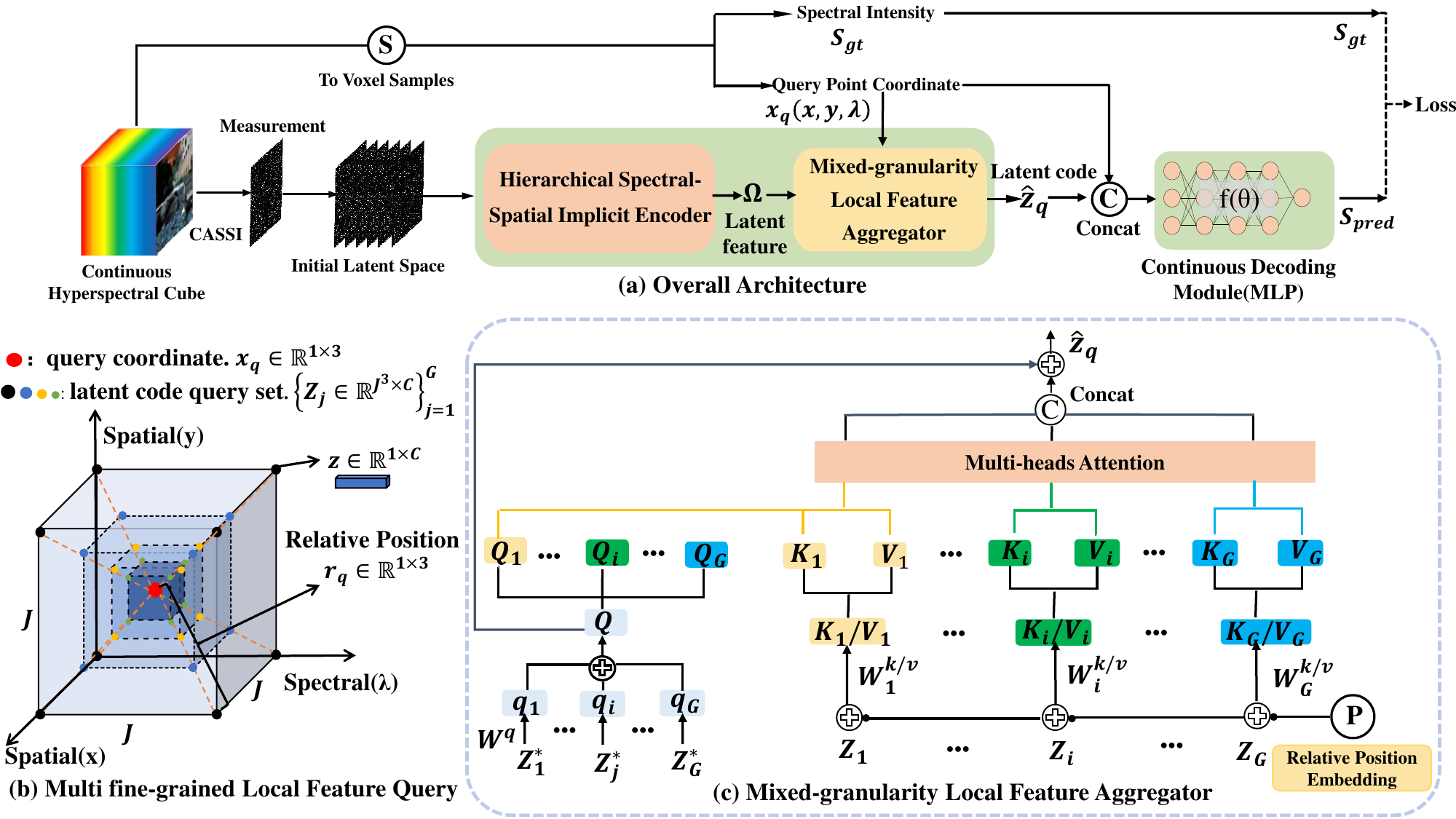}%
    \caption{(a) Overall network architecture of the proposed Mixed-Granularity Implicit Representation, featuring a Hierarchical Spectral-Spatial Implicit Encoder, a Mixed-Granularity Local Feature Aggregator, and an MLP decoder. (b) Multi-granularity Local Feature Query: The red dot represents the query point, with the dots in four distinct colors denoting four different fine-grained features. (c) Illustration of the proposed Mixed-Granularity Local Feature Aggregator.}
   \label{fig3}
\end{figure*}

\noindent\textbf{Implicit Nerual Representation.}
We revisit the concept of implicit neural representation for 2D images, emphasizing its capability for continuous representation. The fundamental principle involves using a shared decoding function across all images to ensure seamless representation~\cite{chen2021learning, lee2022local}.

The image $\boldsymbol I \in \mathbb{R}^{\rm H \times \rm W \times 3} $. The encoder ${\boldsymbol E}_{\varphi}$ extracts latent codes $\boldsymbol Z$ of size $\rm H \times \rm W \times \rm C $, matching the height and width of $\boldsymbol I$, where $\rm C$ denotes the feature dimension:
\begin{align}
& \boldsymbol Z = {\boldsymbol E}_{\varphi}(\boldsymbol I). \label{eq:4}
\end{align}
Coordinates of a query point $x_q$ are input into a decoder, which normalizes them to the interval $[-1,1]$, aligning the predicted image's coordinate domain with that of the original image. The value at the query point is computed as:
\begin{align}
& v_q = \boldsymbol{f}_\theta(z, x_q-x), \label{eq:5}
\end{align}
where $v_q$ is the value at the query pixel $q$, $x_q-x$ represents the relative coordinates, and $\boldsymbol{f}_\theta(\cdot)$ is the decoding function, shared across all images and typically parameterized by a decoder.

The reconstructed image $\boldsymbol{\hat I} \in \mathbb{R}^{\rm H \times \rm W \times 3}$ is computed as follows:
\begin{align}
&\boldsymbol{\hat I}(x_q) = \sum\limits_{i \in {\rm N_q}}w_{q,i}v_{q,i}, \label{eq:6}
\end{align}
where ${\rm N}_q$ is the set of neighboring points around $x_q$ in the original image, and $w_{q,i}$ are the weighting coefficients. Typically, ${\rm N}_q$ includes the $4$ (or $9$) closest points to the predicted pixel.

\subsection {Model Architecture}
To facilitate spectral-spatial continuity in hyperspectral image reconstruction, we introduce the Mixed-Granularity Implicit Representation (MGIR), illustrated in~\cref{fig3}(a). The MGIR framework comprises a Hierarchical Spectral-Spatial Implicit Encoder (HSSIE), a Mixed-Granularity Local Feature Aggregator (MGLFA), and a coordinate-based decoder.

The CASSI system captures a continuous hyperspectral cube and compresses it into a 2D measurement space, denoted as $\boldsymbol M \in \mathbb{R}^{\rm H \times \rm W}$. We extend $\boldsymbol M$ to the initial latent space $\boldsymbol M_0 \in \mathbb{R}^{\rm H \times \rm W \times \rm D \times 1}$, with dimensions representing height, width, and the spectral extent, respectively. 
The Hierarchical Spectral-Spatial Implicit Encoder processes $\boldsymbol M_0$ using a pyramidal structure to efficiently learn hierarchical spectral-spatial representations:
\begin{align}
 \boldsymbol{\Omega} = \boldsymbol{f}_E(\boldsymbol M_0), \label{eq:7}
\end{align}
where $\boldsymbol{f}_E(\cdot)$ denotes the encoding function with learnable parameter $E$.
The Mixed-granularity Local Feature Aggregator refines the latent feature $\boldsymbol{\Omega}$ into latent codes $\boldsymbol Z$, capturing multi-granularity spatial-spectral characteristics:
\begin{align}
\boldsymbol Z = \boldsymbol{f}_A(\boldsymbol{\Omega}), \label{eq:8}
\end{align}
where $\boldsymbol{f}_A(\cdot)$ represents the aggregation function, parameterized by $A$. Finally, the reconstructed hyperspectral image $\boldsymbol {\hat{L}} \in $ is produced by a coordinate-based decoder that integrates the latent codes $\boldsymbol Z$ with spatial-spectral coordinates $\boldsymbol X$:
\begin{align}
\boldsymbol {\hat{L}} = \boldsymbol{f}_{\varphi}(\mathop{\rm{Concat}}(\boldsymbol Z,\boldsymbol X)), \label{eq:9}
\end{align}
where $\mathop{\rm{Concat}}$ is the concatenation of vectors, and $\boldsymbol{f}_{\varphi}(\cdot)$ is a mapping function learned by the decoder, parameterized by $\varphi$.

\begin{figure*}[t]
  \centering
  \includegraphics[width=0.9\linewidth]{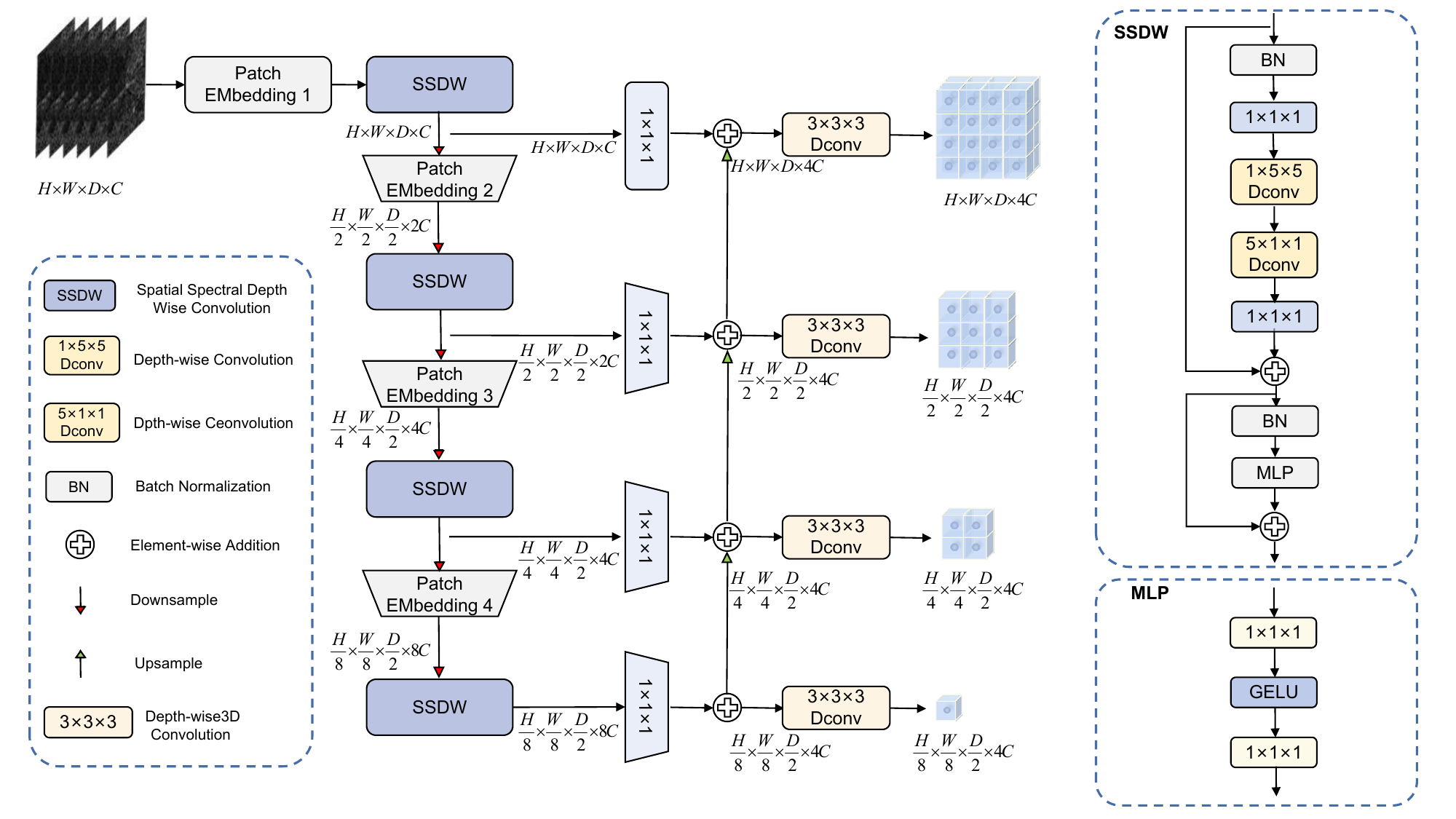}
  \caption{Overall architecture of the Hierarchical Spectral-Spatial Implicit Encoder (HSSIE): This encoder features a hierarchical structure composed of four Spectral-Spatial Depthwise Convolution modules (SSDW) and incorporates patch embeddings.}
   \label{fig4}
\end{figure*}

\subsection{Hierarchical Spectral-spatial Implicit Encoder}
The Hierarchical Spectral-Spatial Implicit Encoder (HSSIE) is designed for efficient and effective learning of spectral-spatial representations. It addresses the limitations of 2D backbones and the high computational load of existing 3D models by offering a computationally efficient alternative that provides a comprehensive feature set across multiple layers. This is achieved by employing spectral-spatial depthwise convolution modules arranged in a layered pyramid structure to construct the encoder.

The encoder processes the initial latent space, $\boldsymbol{M}_0$, by applying a convolution through patch embedding to extract low-level features, avoiding the traditional flattening and projection of a linear layer. This results in features with dimensions $\rm H \times \rm W \times \rm D \times \rm C$, where $\rm C$ represents the feature dimension. Our network architecture, illustrated in~\cref{fig4}, consists of four stages. Each stage integrates a patch embedding module and spectral-spatial depthwise convolution modules to produce a multi-scale latent feature.

\noindent\textbf{Patch Embedding.}
The patch embedding module utilizes 3D convolution and Layer Normalization to progressively halve the spatial-spectral dimensions while doubling the feature dimension. In the $i$-th stage, the output of the embedding is defined as:
\begin{align}
 {\boldsymbol{\Omega}_i^{'}} = \boldsymbol{f}_{PE}(\boldsymbol{\Omega}_{i-1}), \qquad{i=1,2,3,4} \label{eq:10}
\end{align}
where $\boldsymbol{f}_{PE}(\cdot)$ is the mapping function of the patch embedding with learnable parameters PE. $\boldsymbol{{Y_1}^{'}} \in {\mathbb{R}^{ \rm H \times \rm W \times \rm D \times \rm C}}$, and ${\boldsymbol{\Omega}}_i^{'} \in {\mathbb{R}^{\frac{\rm H}{2^{(i-1)}} \times \frac{\rm W}{2^{(i-1)}} \times \frac{\rm D}{2} \times 2^{(i-1)} \rm C }}$ when $i \in \{2,3,4\} $.

\noindent\textbf{Spectral-spatial Depthwise Convolution.}
The Spectral-Spatial Depthwise Convolution (SSDW) is a pivotal component of the encoder, as shown in~\cref{fig4}, designed to efficiently extract joint spatial-spectral features. 

For an input $\boldsymbol{X}_{in}$ to the SSDW, the intermediate output $\boldsymbol{X}_{mid}$ is computed as follows:
\begin{align}
    \boldsymbol X_{mid} = (\boldsymbol{f}_{1}(\text{DSC}(\boldsymbol{f}_{2}(\boldsymbol X_{in}))))+\boldsymbol X_{in}, \label{eq:12}
\end{align}
where $\boldsymbol{f}_{1}(\cdot)$ and $\boldsymbol{f}_{2}(\cdot)$ are 3D convolutions with a kernel size of $1 \times 1 \times 1$. 
The \text{DSC} refers to two Depthwise Separable Convolutions, complemented by a custom-designed MLP. This approach divides the convolutional process into depthwise convolutions, which handle either spatial or spectral features, followed by pointwise convolutions that integrate these features. Specifically, a $1 \times 5 \times 5$ depthwise convolution initially captures spatial features, and a $5 \times 1 \times 1$ depthwise convolution subsequently focuses on spectral features, followed by a $1 \times 1$ convolution for feature integration. This modular design allows for independent processing of spatial and spectral features, significantly reducing the number of parameters and improving computational efficiency without compromising performance.

The output $\boldsymbol{X}_{out}$ from the SSDW is generated by an MLP:
\begin{align}
  \boldsymbol X_{out} = {\boldsymbol{f}_{\theta}}(\boldsymbol X_{mid})+\boldsymbol X_{mid}, \label{eq:13}
\end{align}
where $\boldsymbol{f}_{\theta}(\cdot)$ is a mapping function of the MLP with learnable parameters $\theta$.

\noindent\textbf{Detailed Structure.}
Our hierarchical encoder features a pyramidal structure to optimize latent feature processing in HSIs. Each stage starts by converting the input into an initial feature map via patch embedding, followed by spatial-spectral feature extraction using a SSDW module. Initially, a $1 \times 1 \times 1$ convolution prepares the feature map for integration with upsampled features from subsequent layers. Concurrently, another branch progresses through additional patch embedding and SSDW, preparing inputs for the next stage. This recursive process of branching and aggregation repeats across each level, culminating in the fourth stage. Through upsampling, feature maps from all stages are combined, producing four distinct scales of feature maps that enhance the model's ability to capture and represent multi-scale features.

\noindent\textbf{Computational Cost Analysis of SSDW.}
We conduct a comparative analysis of the computational and memory costs of our SSDW block against global multi-headed self-attention (G-MSA)\cite{dosovitskiy2020image} and local window-based self-attention (W-MSA)\cite{liu2021swin}, as expressed by:
\begin{equation}
    \begin{split}
   O({\text{W-MSA)}} = 4\rm H \rm W \rm D{\rm C^2} + 2{\rm M^3}\rm H \rm W \rm D \rm C,\\
   O({\text{G-WSA) = 4}}\rm H \rm W \rm D{\rm C^2} + 2{(\rm H \rm W \rm D)^2}\rm C,\\
   O({\text{SSDW) = (}}{\rm M^2} + \rm M)\rm H \rm W \rm D \rm C + 2\rm H \rm W \rm D{\rm C^2},
   \end{split}
   \label{computional}
\end{equation}
where $\rm H$, $\rm W$, $\rm D$, $\rm M$ denote the height, width, number of spectral bands, and the convolution kernel size or attention window size, respectively. The computational complexity of SSDW is substantially lower than that of G-MSA and W-MSA.

\subsection{Mixed-granularity Local Feature Aggregator}
The Mixed-Granularity Local Feature Aggregator (MGLFA) compiles latent codes of varying scales from the encoder using a multi-head attention mechanism, effectively capturing multi-scale and multi-granularity spectral-spatial features.

As illustrated in~\cref{fig3}, the MGLFA module begins by accepting input coordinates and utilizing multi-granularity features produced by the encoder to generate queries $\bm Q$, keys $\bm K$, and values $\bm V$. These elements, each representing different granularities, are directed into $\rm G$ groups-set to 4 in our model-where each group operates independently at distinct granularity levels to achieve effective multi-granularity local feature aggregation. Simultaneously, we incorporate relative position encoding to bolster the model's ability to leverage positional information across the various head groups.

In~\cref{fig3}(b), given a query voxel coordinate ${\boldsymbol{x}_q} \in {\mathbb{R}^{1\times3}}$ from $\boldsymbol L$, each query point is situated within a $ \rm{J}\times \rm{J}\times \rm{J} $ neighboring window of multi-scale latent features $\{{\boldsymbol{\Omega_i}\}}_{i=1}^{\rm G}$. For each $\boldsymbol{\Omega}$, we query latent codes within this neighboring window and organize them into clusters, denoted as $\{ {\boldsymbol Z_j} \in {\mathbb{R}^{{\rm{J}^3} \times \rm{C} }}\}_{j = 1}^{\rm G}$.

\noindent\textbf{Gathering $\bm Q$, $\bm K$, $\bm V$.} 
The query voxel coordinate $\boldsymbol x_q \in {\mathbb{R}^{1\times3}}$ corresponds to a latent code set. We first perform trilinear interpolation at each level of the implicit space to access multi-scale context latent codes $\{ \boldsymbol Z_j^* \in {\mathbb{R}^{1 \times \rm{C}}}\} _{j = 1}^{\rm G}$. The queries $\bm{q}$ are then obtained as follows:
 \begin{equation}
     \boldsymbol q_j = \sum\limits_{j = 1}^{\rm G} {\boldsymbol Z_j^*  \boldsymbol W_j^Q},\qquad{j = 1,...,{\rm G}}
    \label{eq:15}
 \end{equation}
where ${\boldsymbol W_{j}^Q \in {\mathbb{R}^{\rm C \times \rm C}}}$ are learned weight matrices of linear layers. The query vector $\bm{Q}$ is obtained by element-wise addition of the query vectors $\boldsymbol{q_j}$.

To achieve granularity-aware attention learning, we divide multiple attention heads into $ \rm G $ groups, each responsible for attending to information within a particular range. Consequently,  we split $ \bm Q $ into $ \boldsymbol Q_j $, $ j = 1, \ldots, \rm G $. To gather context across different receptive field scales, the keys and values are computed as follows:
 \begin{align}
    &{\bm K_j} = {\boldsymbol Z_j}\boldsymbol W_j^ K,\qquad{j = 1,...,\rm G}\\
    &{\bm V_j} = {\boldsymbol Z_j}\boldsymbol W_j^V,\qquad{j = 1,...,\rm G}.
    \label{eq:16}
\end{align}
where $\boldsymbol W_j^K, \boldsymbol W_j^V \in {\mathbb{R}^{\rm C \times \rm C}}$ are learnable parameters of linear projection in the $j$-th scale. The outputs of all groups are merged to obtain the final mixed-granularity features:
\begin{equation}
    {\boldsymbol{\hat {Z}}_q} = \mathop{\rm{Concat}}_{j = 1}^{\rm G}({\boldsymbol{MHA}}({\bm Q_j},\boldsymbol{RPE}(\bm K_j),\boldsymbol{RPE} (\bm V_j))) + {\bm Q},
\label{eq:17}
\end{equation}
where $\boldsymbol {MHA}(\cdot)$ denotes multi-headed attention~\cite{vaswani2017attention, voita2019analyzing}, and $\boldsymbol {RPE}(\cdot)$ represents the relative position encoding, which will be described in detail later.

\noindent\textbf{Relative Position Encoding.} 
To enhance the positional information within the Mixed-granularity Local Feature Aggregator, we have integrated relative position embeddings into both the keys and values. This approach is intended to refine the distinction between the characteristics of adjacent voxels, ensuring the preservation of spatial relationships throughout the network's architecture. Specifically, voxel coordinates are mapped to a higher-dimensional space using high-frequency functions, and distance information is embedded into the keys $\bm{K}$ and values $\bm{V}$. The mapping function is defined as:
\begin{equation}
 \boldsymbol{RPE}(x) = [{\rm{cos}}({\omega_i}x), {\rm{sin}}({\omega_i}x)]^\top,
    \label{eq:18}
\end{equation}
where $\omega_i$ initially takes the values $2e^i$, $i \in \{1, 2,...,\rm{R}\}$. These frequency parameters are further refined during training. The sinusoidal functions, serving as encoding mechanisms, operate on coordinates at varied frequencies and independently combine them, thus transforming the space from $\boldsymbol X \in {\mathbb{R}^3}$ to $\boldsymbol X_p \in {\mathbb{R}^{\rm{6R}}}$. A linear layer then projects $\boldsymbol X_p$ onto a $C$-dimensional position embedding to align with the dimensions of the input key/value pairs. The embedded positional information is subsequently incorporated into $\bm{K}$ and $\bm{V}$ as follows:
\begin{align}
    &{\bm K_j} = {\bm K_j}+\boldsymbol X_p,\\
    &{\bm V_j} = {\bm V_j}+\boldsymbol X_p.
    \label{eq:19}
\end{align}
Thus, relative position encoding is adaptively adjusted across different scales to guide mixed-granularity feature learning.

\begin{table*}[t]
\centering
\caption{Quantitative results for continuous spectral reconstruction on the ICVL dataset.}
\renewcommand\arraystretch{1.1}
\setlength{\tabcolsep}{8.0pt}
\begin{tabular}{l|ccc|ccc|ccc|ccc}
\toprule
\multirow{2}{*}{Method} & \multicolumn{3}{c|}{31 Bands} & \multicolumn{3}{c|}{62 Bands} & \multicolumn{3}{c|}{124 Bands} & \multicolumn{3}{c}{241 Bands} \\ \cline{2-13}
 & \multicolumn{1}{c}{RMSE} & \multicolumn{1}{c}{PSNR} & \multicolumn{1}{c|}{SAM} & \multicolumn{1}{c}{RMSE} & \multicolumn{1}{c}{PSNR} & \multicolumn{1}{c|}{ SAM} & \multicolumn{1}{c}{RMSE } & \multicolumn{1}{c}{PSNR} & \multicolumn{1}{c|}{SAM} & \multicolumn{1}{c}{RMSE} & \multicolumn{1}{c}{PSNR} & \multicolumn{1}{c}{SAM} \\ 
 \midrule
3D U-Net~\cite{esmaeilzadeh2020meshfreeflownet} & 0.035 & 31.96 & 0.102 & 0.044 & 30.05 & 0.149 & 0.044 & 29.97 & 0.148 & 0.044 & 29.90 & 0.147 \\
MST~\cite{cai2022mask} & 0.014 & 39.22 & 0.052 & 0.022 & 36.24 & 0.122 & 0.026 & 35.89 & 0.132 & 0.029 & 34.94 & 0.144 \\
CST~\cite{cst} & 0.013 & 39.63 & 0.048 & 0.024 & 36.31 & 0.133 & 0.029 & 35.76 & 0.151 & 0.029 & 35.60 & 0.151 \\
MGIR (Ours) &\textbf{0.011}  & \textbf{40.21} &\textbf{0.041}  &\textbf{0.015}  &\textbf{38.51} &\textbf{0.065} &\textbf{0.015} &\textbf{38.67} &\textbf{0.058} &\textbf{0.014} &\textbf{39.07} &\textbf{0.054}   \\ 
\bottomrule
\end{tabular}
\label{tab:Continuous Spectral Reconstruction}
\end{table*}

\subsection{Hyperspectral Image Implicit Neural Function}
In the context of Implicit Neural Representation, we posit that the latent codes, shaped as $\rm H \times \rm W \times \rm D$, are uniformly distributed throughout the 3D spectral-spatial domain, with each code assigned a corresponding 3D coordinate. To model the spectral-spatial intensities, we employ a vector-intensity function ${\boldsymbol f_\theta }$, parameterized by a learnable $\theta$. This function takes as input the coordinates $(x,y,\lambda)$ along with the latent codes and outputs the respective spectral-spatial intensity. The initial predicted spectral intensity values are defined as:
\begin{equation}
  \boldsymbol{\hat{L}} = {\boldsymbol f_\theta}(\boldsymbol{Z},\boldsymbol{X}),
\end{equation}
where $\boldsymbol{Z}$ represents the latent codes refined by the implicit encoder and the mixed-granularity feature aggregator; $\boldsymbol{X}$ denotes the spectral-spatial coordinates associated with $\boldsymbol L$; and $\boldsymbol{\hat{L}}$ indicates the predicted spectral intensity values. The decoding function, ${\boldsymbol f_\theta}$, is common across all hyperspectral images and is implemented using a MLP.

The Root Mean Square Error (RMSE) is selected as the loss function to optimize our network for accurate predictions. It is defined as:
 \begin{equation}
\begin{aligned} \label{eq:21}
    \mathcal{L}_\text{RMSE} = \sqrt {\frac{1}{\rm N}\sum\nolimits_{i = 1}^{\rm N} {{{({\boldsymbol {L}^{(i)}} - {\boldsymbol{\hat L}^{(i)}})}^2}}},
\end{aligned}
\end{equation}
where $\boldsymbol{\hat L}^{(i)}$ and $\boldsymbol{L}^{(i)}$ represent the predicted and ground truth values of the ${i^\text{th}}\ (i = 1,2,...,\rm {N})$ voxel in the hyperspectral image, respectively, and $\rm N$ denotes the total number of voxels.

\section{Experiments}
\subsection{Implementation Details}\label{implementation}
The Mixed-granularity Local Feature Aggregator incorporates four attention heads. In our encoder, the depths of the SSDW stages are set to (2,2,4,4). The window size for neighboring multi-scale latent feature extraction is established at $2 \times 2 \times 2$. The dimension of the coordinate mapping in Relative Position Encoding is configured to 120. The model is trained using the Adam optimizer~\cite{kingma2014adam} for a total of 300 epochs and a batch size of 4. The initial learning rate is $4 \times 10^{-4}$.
Experiments are executed on two NVIDIA RTX 3090 GPUs, paired with an Intel XEON Gold 5218R CPU.

We evaluate our model using the ICVL~\cite{arad_and_ben_shahar_2016_ECCV}, CAVE~\cite{Huang_2017}, and KAIST~\cite{choi2017high} datasets. In the spectral domain, we select bands within the visible range of 400 to 700 $\rm nm$ and preprocess the hyperspectral images by randomly cropping them to a spatial size of $256 \times 256$. In line with configurations reported in~\cite{wang2021tsa, cai2022mask, hu2022hdnet}, we set the dispersion shift $\rm d$ in the CASSI system to 2, yielding a measurement size of $256 \times 310$. Data augmentation includes random horizontal and vertical flips.

We utilize the Peak Signal-to-noise Ratio (PSNR), Structural Similarity Index Measure (SSIM)~\cite{DBLP:journals/corr/Larkin15}, Root Mean Square Error (RMSE), and Spectral Angle Mapping (SAM)~\cite{kruse1993spectral} as evaluation metrics.

\begin{figure}[t]
  \centering
  \includegraphics[width=1.0\linewidth]{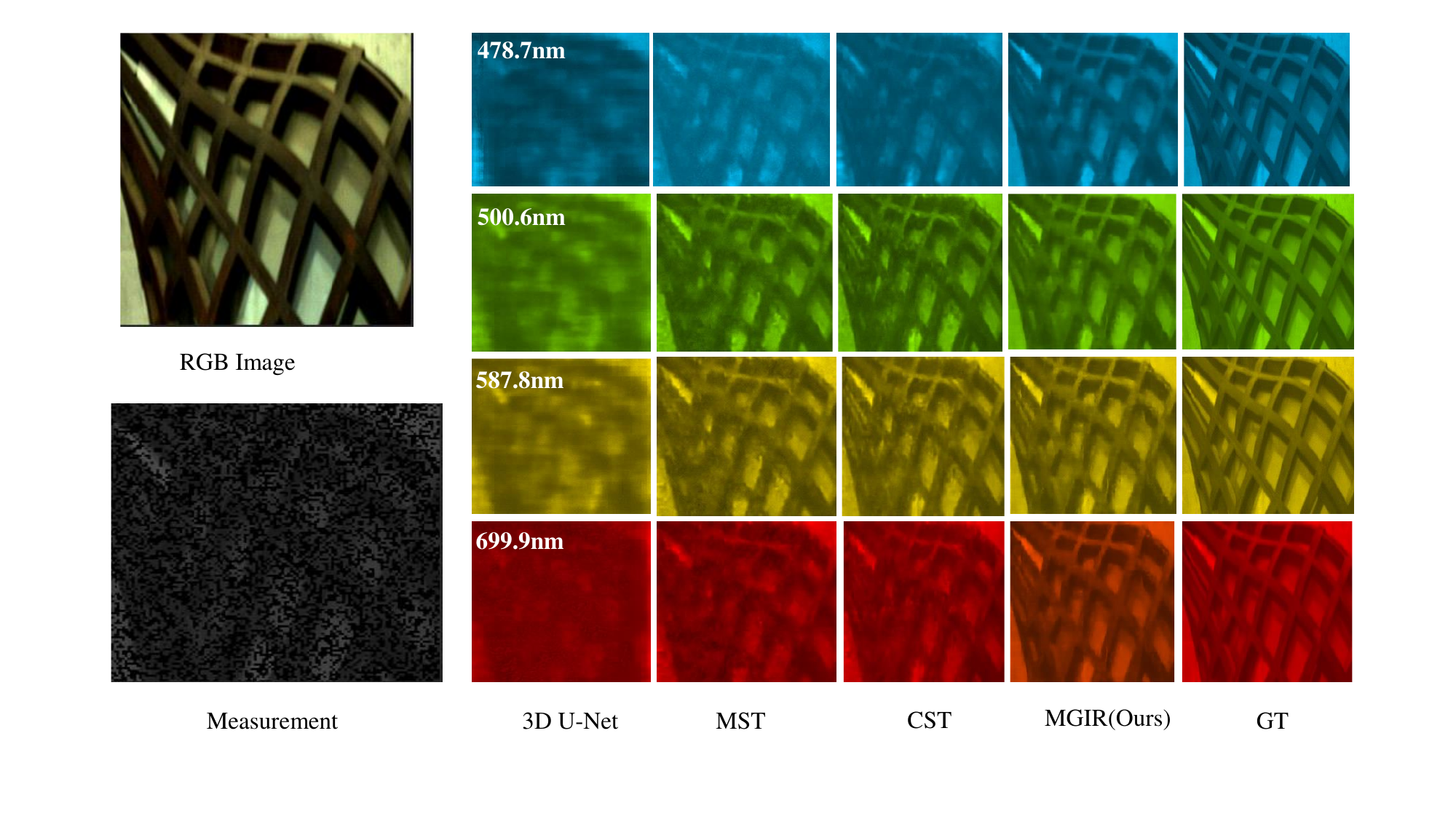}
    \caption{Visualization of reconstructed results (top) from the ICVL dataset, featuring a selection of 4 out of 241 spectral bands.}
   \label{reconstruction visual result of 241 bands}
\end{figure} 

\subsection{Continuous Spectral Reconstruction on ICVL}
\noindent\textbf{Data and Setups.} 
The ICVL dataset comprises 201 images, each with a spatial resolution of $1392 \times 1300$ and covering 519 spectral bands ranging from 400 to 1000 $\rm nm$ at approximately 1.25 $\rm nm$ intervals. We select 241 spectral bands from the full range that fall within the visible spectrum. After eliminating images with excessive scene repetition, we are left with a refined dataset of 95 images. From this curated collection, we randomly select 85 images to form our training set, reserving the remaining 10 for testing purposes.

During the training phase, we focus on compressive spectral reconstruction using a subset of 31 spectral bands, chosen at 10 $\rm nm$ intervals from 241 bands. We configure the number of spectral bands to be multiples of 31 to facilitate comparison and reconstruction.
It is important to highlight that during the training process, no spectral bands other than the selected 31 are utilized. In our testing phase, we evaluate the performance of our MGIR method by compressing and reconstructing 31, 62, 124, and 241 spectral bands. For other methods, we perform tests on 31 bands at a time and aggregate the results to achieve the desired number of spectral bands.

\begin{figure}[t]
  \centering
  \includegraphics[width=0.95\linewidth]{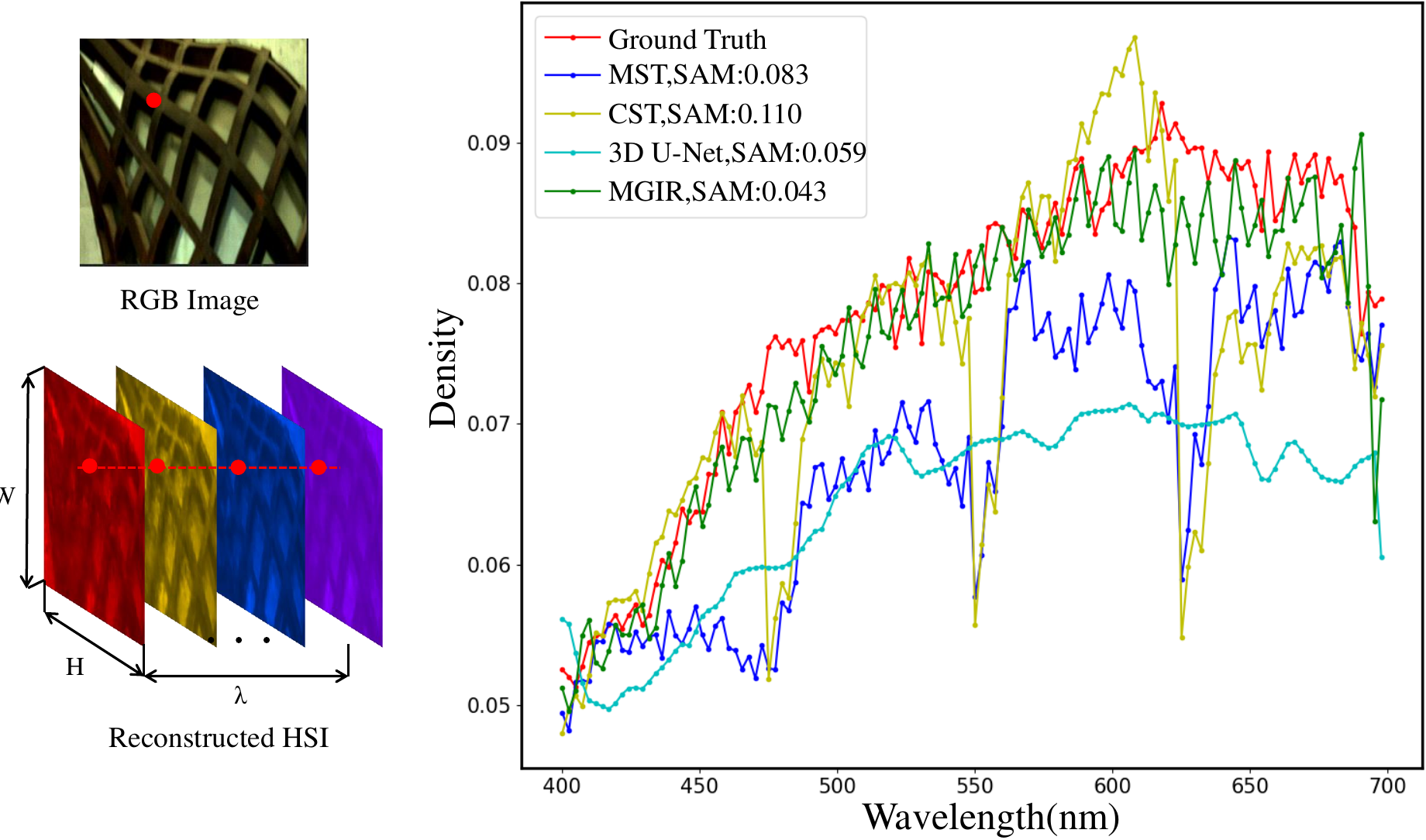}
    \caption{Reconstruction of the spectral density curve for 124 spectral bands at the location marked by a red dot on the RGB image. The spectral curves from our MGIR model show the highest correlation with the ground truth.}
   \label{reconstruction spectral curve of 124 bands}
\end{figure} 

\begin{table*}[t]
\centering
\caption{Quantitative results of continuous spatial reconstruction. Spatial CR: spatial compression ratio.}
\renewcommand\arraystretch{1.1}
\setlength{\tabcolsep}{6.0pt}
\begin{tabular}{c|c|ccc|ccc|ccc|ccc}
\toprule
\multirow{2}{*}{Method} & \multirow{2}{*}{Spatial CR} & \multicolumn{3}{c|}{MST + Trilinear} & \multicolumn{3}{c|}{MST + LIIF} & \multicolumn{3}{c|}{CST + LIIF} & \multicolumn{3}{c}{MGIR (Ours)} \\ \cline{3-14}
  & & \multicolumn{1}{c}{PSNR} & \multicolumn{1}{c}{SSIM} & \multicolumn{1}{c|}{SAM} & \multicolumn{1}{c}{PSNR} & \multicolumn{1}{c}{SSIM} & \multicolumn{1}{c|}{ SAM} & \multicolumn{1}{c}{PSNR} & \multicolumn{1}{c}{SSIM} & \multicolumn{1}{c|}{SAM} & \multicolumn{1}{c}{PSNR} & \multicolumn{1}{c}{SSIM} & \multicolumn{1}{c}{SAM} \\  
 \midrule
\multirow{2}{*}{In-distribution} & $\times 2$ & 34.19 & 0.926 & 0.069 & 34.97 & 0.934 & 0.067 & 35.61 & 0.932 & 0.067 & \bf{36.49} & \bf{0.944} & \bf{0.061}\\
 & $\times 4$ & 29.80 & 0.884 & 0.104 & 30.59 & 0.901 & 0.107 & 31.19 & 0.904& 0.102 & \bf{33.29} &\bf{ 0.917} & \bf{0.079} \\
 \midrule
\multirow{2}{*}{Out-of-distribution} & $\times 8$ & 27.02 & 0.847 & 0.123 & 27.86 & 0.862 & 0.115 & - & - & - & \bf{30.98} & \bf{0.894}& \bf{0.095} \\
 & $\times 16$ & 26.30 & 0.834 & 0.163 & 27.23 & 0.848 & 0.142 & - & - & - &\bf{28.96} &\bf{0.872} &\bf{0.112} \\ 
 \bottomrule
\end{tabular}
\label{spatial result}
\end{table*}

\begin{figure*}[t]
  \centering
  \includegraphics[width=0.95\linewidth]{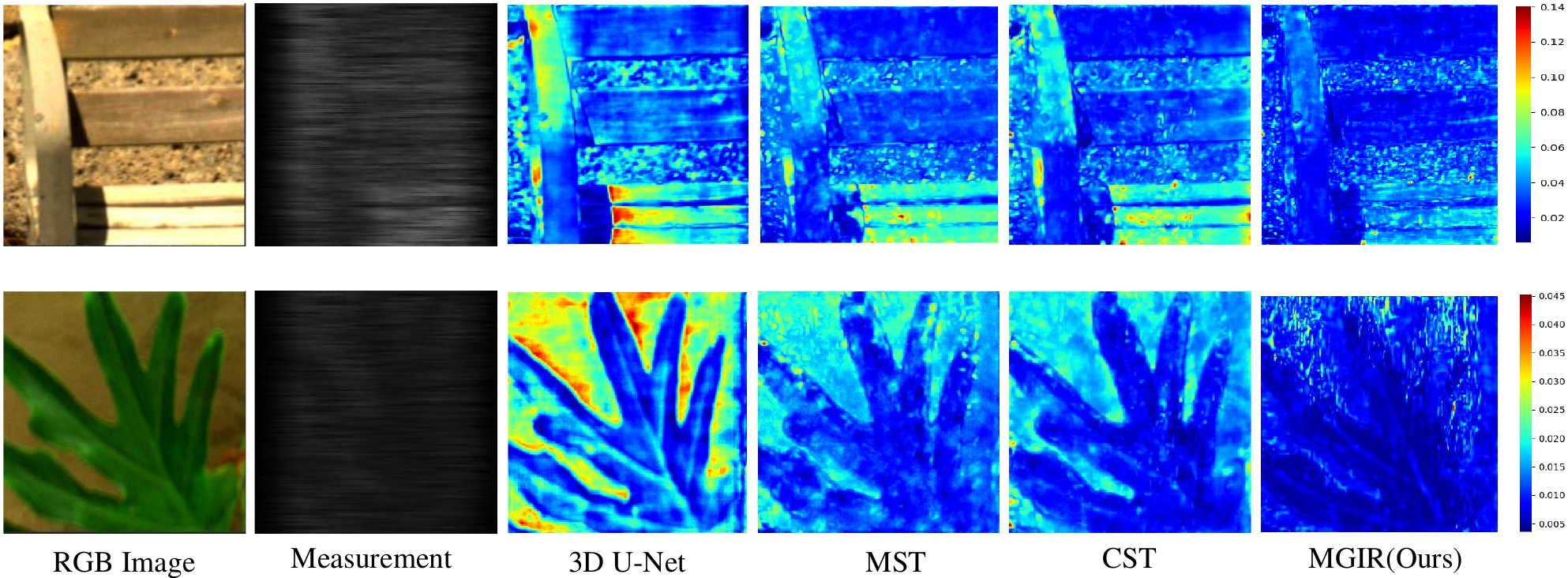}
   \caption{Visualization of error maps for various spectral reconstruction methods on the ICVL dataset, encompassing 124 spectral bands. A bluer color indicates a smaller reconstruction error, while a redder color signifies a larger error.}
   \label{errop 124}
\end{figure*}

\begin{table*}
\centering
\caption{Quantitative results of spatial reconstruction of 10 scenes on KAIST. PSNR/SSIM (the first/second line) are reported.}
\label{cave_data}
\renewcommand\arraystretch{1.1}
\setlength{\tabcolsep}{4.0pt}
\begin{tabular}{ l | c  c | c  c  c  c  c  c  c  c  c  c  | c c}
\toprule
Methods & \#Params & FLOPs & Scene1 & Scene2 & Scene3 & Scene4 & Scene5 & Scene6 & Scene7 & Scene8 & Scene9 & Scene10 & Average \\
\midrule
\multirow{2}{*}{TSA-Net~\cite{meng2020end}}
&\multirow{2}{*}{44.25 $\rm M$}  & \multirow{2}{*}{110.06 $\rm G$}  & 32.03  & 31.00 & 32.25 & 39.19 & 29.39 & 31.44 & 30.32 & 29.35& 30.01 & 29.59 & 31.46\\
& & & 0.892 & 0.858 & 0.915 & 0.953 & 0.884 & 0.908 & 0.878 & 0.888 & 0.890 & 0.874 & 0.894\\
\hline

\multirow{2}{*}{GAP-Net~\cite{meng2023deep}}
&\multirow{2}{*}{4.27 $\rm M$}  & \multirow{2}{*}{78.58 $\rm G$}  & 33.74 &33.26&  34.28  &41.03 & 31.44 & 32.40 & 32.27  &30.46 & 33.51 & 30.24 &33.26\\
& & &0.911 &0.900 &0.929 &0.967& 0.919 &0.925&  0.902  &0.905 & 0.915 & 0.895  &0.917\\
\hline

\multirow{2}{*}{HDNet~\cite{9879297}}
&\multirow{2}{*}{2.37 $\rm M$}  & \multirow{2}{*}{154.76 $\rm G$}  & 35.14 & 35.67 & 36.03 & 42.30 & 32.69 & 34.46 & 33.67 & 32.48 & 34.89 & 32.38 & 34.97\\
& & &0.935 & 0.940 & 0.943 & 0.969 & 0.946 & 0.952 & 0.926 & 0.941 & 0.942 & 0.937 & 0.943\\
\hline

\multirow{2}{*}{MST-S~\cite{mst}}
&\multirow{2}{*}{0.93 $\rm M$} & \multirow{2}{*}{12.96 $\rm G$} &34.71 & 34.45& 35.32 & 41.50 &31.90 & 33.85 & 32.69 & 31.69 & 34.67 & 31.82 &34.26\\
& & & 0.930 & 0.925 &0.943&0.967 & 0.933 & 0.943 & 0.911 &0.933 & 0.939 &0.926& 0.935\\
\hline
\multirow{2}{*}{MST-M~\cite{mst}}
&\multirow{2}{*}{1.50 $\rm M$} & \multirow{2}{*}{18.07 $\rm G$}&  35.15 & 35.19& 36.26 & 42.48 & 32.49 & 34.28 & 33.29 & 32.40 & 35.35 & 32.53 &34.94\\
& & & 0.937 & 0.935 & 0.950 & 0.973 & 0.943 & 0.948 &0.921 & 0.943 & 0.942 & 0.935 &0.943\\
\hline
\multirow{2}{*}{MST-L~\cite{mst}}
&\multirow{2}{*}{2.03 $\rm M$} & \multirow{2}{*}{28.15 $\rm G$} & 35.40 & \bf{35.87} & 36.51 & 42.27 & 32.77 & 34.80 & 33.66 & 32.67 & 35.39 & 32.50 & 35.18\\
& & & \bf{0.941} & \bf{0.944} & 0.953 & 0.973 & 0.947 & 0.955 &0.925 & 0.948 & 0.949 &0.941 &0.948&\\
\hline             
\multirow{2}{*}{CST-S~\cite{cst}}
&\multirow{2}{*}{1.20 $\rm M$} & \multirow{2}{*}{11.67 $\rm G$}&  34.78 & 34.81 & 35.42 & 41.84 & 32.29 & 34.49 & 33.47 & 32.89 & 34.96 & 32.14 & 34.71\\
& & & 0.930 & 0.931 & 0.944 & 0.967 & 0.939 & 0.949 & 0.922 & 0.945 & 0.944 & 0.932 &0.940\\
\hline
\multirow{2}{*}{CST-M~\cite{cst}}
&\multirow{2}{*}{1.36 $\rm M$} & \multirow{2}{*}{16.91 $\rm G$}& 35.16 & 35.60 & 36.57 & 42.29 & 32.82 & \bf{35.15} & 33.85 & 33.52 & 35.28 & \bf{32.84} & 35.31\\
& & & 0.938 & 0.942 & 0.953 & 0.972 &0.948 &\bf{0.956}& 0.927 & 0.952 & 0.946 & 0.940 &0.947\\

\hline
\multirow{2}{*}{MGIR (Ours)}
&\multirow{2}{*}{0.92 $\rm M$} & \multirow{2}{*}{33.05 $\rm G$} &\textbf{35.72}  &35.29  &\textbf{37.82}  &\textbf{43.43}  &\textbf{33.75}  &34.15  &\textbf{34.52}  &\textbf{33.64}  &36.26  &32.34  &\textbf{35.69} \\  
& & &\textbf{0.941}   &0.943   &0.956  &\textbf{0.980}  &\textbf{0.952}  &0.948  &\textbf{0.955}  &\textbf{0.947}  &\textbf{0.951}  & \textbf{0.942}  & \textbf{0.951}
\\ 

\bottomrule
\end{tabular}
\end{table*}

\begin{figure*}[t]
  \centering
  \includegraphics[width=0.90\linewidth]{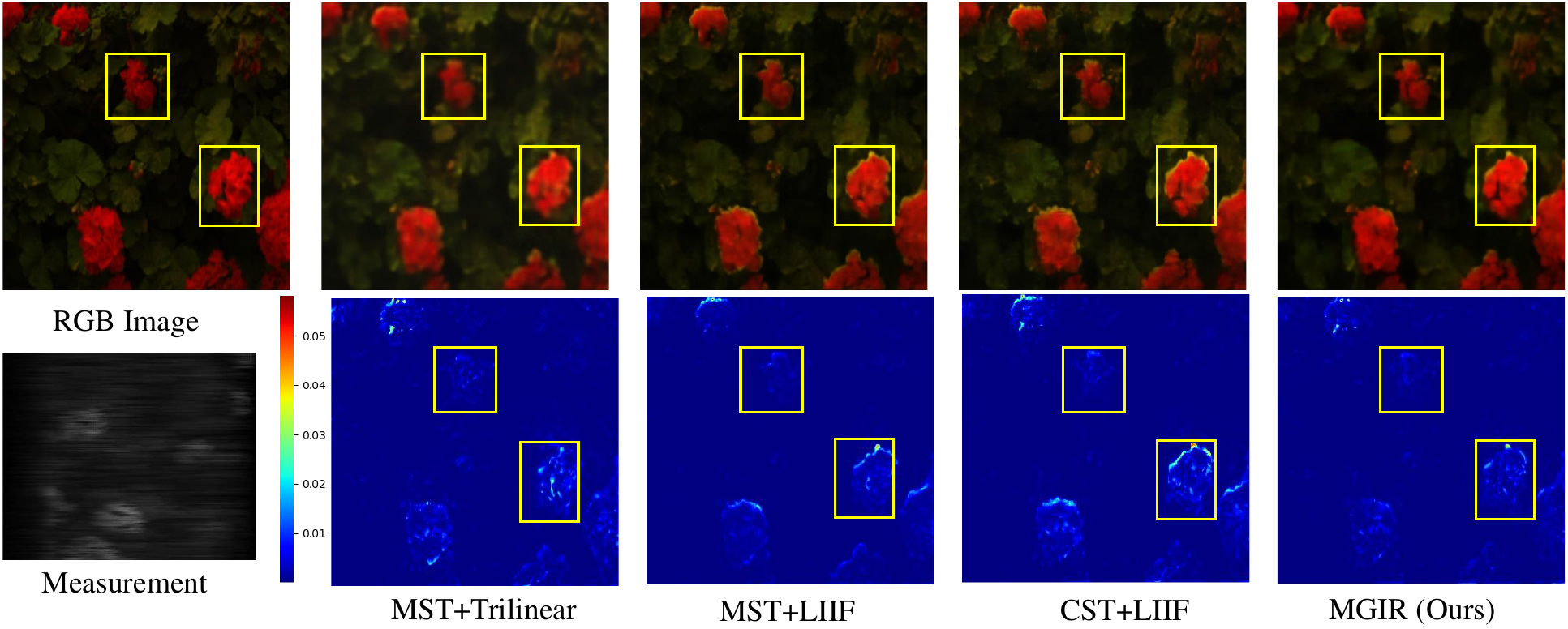}
   \caption{Visualization of continuous spatial reconstruction ($\times 4$) using different methods on the ICVL dataset. The original spatial dimensions are $1024 \times 1024$. We reduce the spatial size of the hyperspectral image by a factor of four using trilinear interpolation, followed by reconstruction to the original dimensions.}
   \vspace{-0.2cm}
   \label{spatial 4 visual result}
\end{figure*}


\noindent\textbf{Quantitative Results.}
We benchmark our method against two discrete reconstruction techniques, MST~\cite{mst} and CST~\cite{cst}, as well as a 3D U-Net-based implicit representation method~\cite{esmaeilzadeh2020meshfreeflownet} for comparison.  
For the standard reconstruction task at 31 bands, MGIR outperforms MST, CST, and 3D U-Net, achieving the highest PSNR of 40.21 $\rm dB$, which is 0.59 $\rm dB$, 0.58 $\rm dB$, and 8.25 $\rm dB$ higher than MST, CST, and 3D U-Net, respectively. MGIR also achieves the lowest RMSE of 0.011 and the best SAM value of 0.041, indicating superior reconstruction accuracy and preservation of spectral features.  
In non-standard reconstruction tasks at 62, 124, and 241 bands, MGIR consistently demonstrates the smallest performance drop. For example, at 241 bands, MGIR achieves a PSNR of 39.07 $\rm dB$, with a minimal RMSE of 0.014 and a SAM value of 0.054. These results highlight the robustness of MGIR in continuous spectral reconstruction, as it not only maintains high accuracy but also preserves spectral feature continuity across all spatial and spectral dimensions.  
Overall, MGIR demonstrates exceptional performance across standard and non-standard band reconstructions.

\cref{fig1}(b) offers a visual comparison. The left vertical axis represents PSNR in $\rm dB$, the right vertical axis is SAM, and the horizontal axis denotes the continuous spectral reconstruction bands. Our MGIR is capable of reconstructing an arbitrary number of spectral bands, including those not present in the training sample. For instance, MGIR can accurately reconstruct 100 and 200 bands, which are not multiples of 31, demonstrating that the number of reconstructed bands by MGIR is unrestricted. In contrast, previous methods like MST and CST cannot achieve arbitrary band reconstruction.

Significantly, while methods such as MST exhibit a decline in reconstruction quality with an increasing number of spectral bands, our method maintains high performance under the same conditions. This underscores the limitation of previous models, which are constrained by the number of reconstructed bands. MGIR, however, provides efficient and adaptable reconstruction across various spectral compression ratios.

\begin{figure*}[t]
  \centering
  \includegraphics[width=0.90\linewidth]{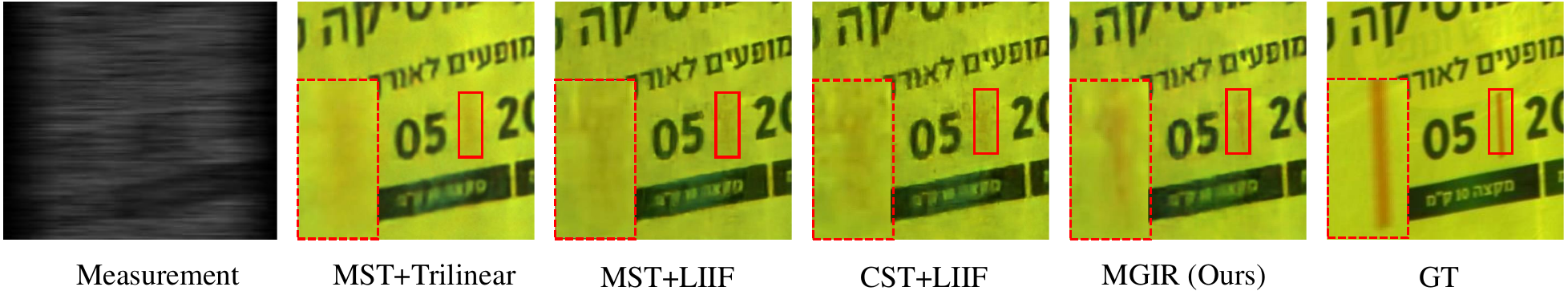}
  \caption{Visualization of continuous spatial reconstruction ($\times 8$) using different methods on the ICVL dataset.}
   \label{spatial 8 visual result}
\end{figure*}

\noindent\textbf{Qualitative Results.}
\cref{reconstruction visual result of 241 bands} presents a visual comparison of hyperspectral image reconstructions, showcasing four distinct spectral channels out of the 241 available. Our approach delineates fine-grained structural elements and textural details with superior clarity. We depict the spectral density curves of reconstructed images in~\cref{reconstruction spectral curve of 124 bands}, covering 124 spectral bands across different methods. The spectral density analysis reveals that our method maintains fidelity to the ground truth, ensuring high spectral accuracy and preserving fine high-frequency spectral details, even at an increased band count of 124.

Additionally, we provide two representative reconstruction error maps between the ground truth and the reconstructed images for 124 spectral channels in~\cref{errop 124}. The results demonstrate that the Mixed-Granularity Implicit Representation outperforms other methods in minimizing reconstruction errors, reflecting superior spatial and spectral reconstruction accuracy.

\begin{figure*}[t]
  \centering
  \includegraphics[width=0.95\linewidth]{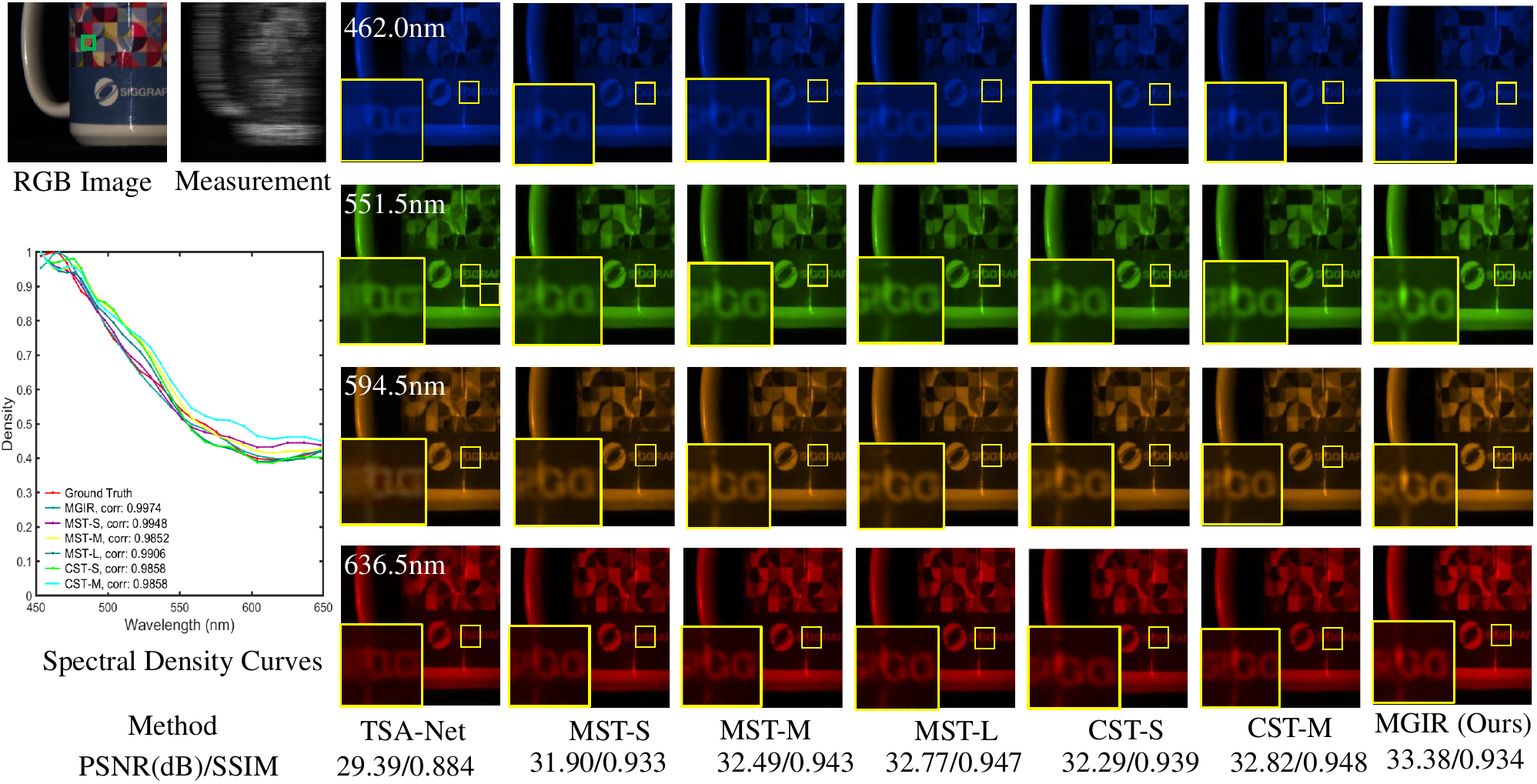}
  \caption{Visualization of reconstructed results for 4 out of 28 spectral channels. MGIR demonstrates precise reconstruction with parameters comparable to three state-of-the-art methods. Spectral curves in the bottom-left are associated with the area marked by the green box on the RGB image.}
   \label{cave_band28}
\end{figure*}

\begin{table*}
\centering
\caption{Comparative analysis of performance, parameters, and FLOPS between W-MSA and G-MSA.}
\setlength{\tabcolsep}{10.1pt}
\label{tab:light}
\begin{tabular}{l| c c c c| c c c c| c c c c}
\toprule
\multirow{2}{*}{Method} & \multicolumn{4}{c|}{W-MSA} & \multicolumn{4}{c|}{G-MSA} & \multicolumn{4}{c}{SSDW (Ours)} \\ \cline{2-13}
& $\times$31 & $\times$62 & $\times$124  & $\times$241 & $\times$31 & $\times$62 & $\times$124 & $\times$241 & $\times$31 & $\times$62 & $\times$124 & $\times$241 \\ 
\midrule[0.7pt]
PSNR & 40.54 & 37.53 & 37.78 & 37.61 & 40.45 & 37.49 & 37.83 & 37.69 & 40.85 & 37.89 & 38.13 & 38.15 \\
SSIM & 0.967 & 0.961 & 0.961  & 0.953 & 0.968 & 0.961 & 0.961 & 0.960 & 0.968 & 0.962 & 0.962 &0.963 \\
SAM & 0.041 & 0.099 & 0.107 &0.164 & 0.040 & 0.089 & 0.086 &0.081 & 0.040 & 0.086 & 0.082 &0.076 \\
\midrule
FLOPs & \multicolumn{4}{c|}{3.78 $\rm G$} & \multicolumn{4}{c|}{3.34 $\rm G$} & \multicolumn{4}{c}{2.94 $\rm G$} \\
\# Params & \multicolumn{4}{c|}{1.98 $\rm M$} & \multicolumn{4}{c|}{2.00 $\rm M$} & \multicolumn{4}{c}{1.66 $\rm M$} \\
\bottomrule
\end{tabular}
\vspace{-0.4cm}
\end{table*}

\subsection{Continuous Spatial Reconstruction on ICVL}
In addition to the reconstruction of continuous spectral bands, our method is capable of reconstructing hyperspectral images at arbitrary spatial resolutions. To substantiate the versatility of our approach in spatial resolution reconstruction, we employ spatial trilinear interpolation in conjunction with the MST as a comparative baseline. Furthermore, we utilize the combination of the 2D implicit representation function LIIF~\cite{chen2021learning} with both the MST and CST reconstruction methods as alternative comparative approaches.

\noindent\textbf{Quantitative Results.} 
\cref{spatial result} presents a quantitative comparison of spatial reconstruction accuracy between our MGIR and the three aforementioned methods, under both in-distribution conditions (spatial compression ratios of $\times 4$, $\times 8$) and out-of-distribution conditions (spatial compression ratios of $\times 8$, $\times 16$). The number of reconstructed spectral bands is consistently set to 31 for this evaluation.

\cref{spatial result} indicates that MGIR maintains the highest spatial accuracy across all tested compression ratios, both within and beyond the distribution range. The MST combined with Trilinear interpolation performs the poorest, while MGIR outperforms MST combined with LIIF by 2.0, 3.2, 3.01, and 1.93 $\rm dB$ at scales of $\times 2$, $\times 4$, $\times 8$, and $\times 16$, respectively. The MST+LIIF method, which employs 2D neural implicit reconstruction, shows only a slight improvement over direct interpolation, suggesting that the 2D continuous representation may not fully encapsulate the multichannel information inherent in hyperspectral images. The significant deterioration in SAM results for both MST+LIIF and CST+LIIF implies that the 2D continuous representation of 3D hyperspectral images is not efficient. In contrast, MGIR leverages a multi-granularity 3D implicit representation, which substantially enhances the accuracy of spatial reconstruction and illustrates the advantages of out-of-distribution generalization.

\noindent\textbf{Visualization under Different Compression Ratios.}
\cref{spatial 4 visual result} illustrates the reconstructed outcomes at a $\times 4$ spatial compression ratio, which aligns with the training distribution. The comparison of the areas highlighted by boxes in~\cref{spatial 4 visual result} reveals that the RGB rendering of the reconstructed hyperspectral images more closely approximates the original RGB image, particularly in capturing the texture details within the flowers. Furthermore, the error map highlights that our method exhibits a reduced overall error compared to several other techniques, particularly notable in the floral regions.

In~\cref{spatial 8 visual result}, we present a qualitative analysis at a $\times 8$ spatial scale, beyond the scope of training. Here, MGIR demonstrates superior reconstruction performance, maintaining high fidelity at this significantly increased scale. Both~\cref{spatial 4 visual result} and~\cref{spatial 8 visual result} affirm the efficacy of our method across scenarios that are both within and beyond the training distribution.

\subsection{{Results on CAVE and KAIST}}
\noindent\textbf{Data and Setups.} 
We also benchmark the performance of MGIR against state-of-the-art methods in reconstructing spectral images across a specific number of spectral bands. The experimental setup replicates that of TSA-Net~\cite{wang2021tsa}. We conduct simulation experiments on the CAVE and KAIST datasets. The CAVE dataset comprises 32 hyperspectral images with a spatial size of $512 \times 512$, while KAIST consists of 30 hyperspectral images with a resolution of $2704 \times 3376$. The spectral range for both the CAVE and ICVL datasets spans from 400 $\rm nm$ to 700 $\rm nm$, segmented into 28 spectral bands. In alignment with methodologies described in~\cite{mst,cst}, we train MGIR using the CAVE dataset and evaluate its performance on 10 scenes from KAIST.

\noindent\textbf{Quantitative Results.}
As shown in~\cref{cave_data}, MGIR achieves an average PSNR of 35.69 $\rm dB$ and SSIM of 0.951, significantly outperforming TSA-Net, GAP-Net, HDNet, CST, and the MST series of methods. Specifically, MGIR surpasses TSA-Net, GAP-Net, MST-S, MST-M, and CST-S by 3.51 $\rm dB$, 1.71 $\rm dB$, 0.71 $\rm dB$, 0.03 $\rm dB$, and 0.26 $\rm dB$ in PSNR, respectively, demonstrating its significant advantage in image quality reconstruction. At the same time, MGIR achieves the best performance in almost every test scenario, further proving its generalization capabilities and reconstruction accuracy across different scenes.
Moreover, despite using a 3D backbone network as the encoder, MGIR has a model parameter count of only 0.92 $\rm M$ and a computational complexity of 33.05 $\rm G$, which is comparable to, or even lower than, conventional 2D CNNs or transformer-based models. This is particularly noteworthy as it means MGIR maintains low computational costs and parameter counts while sustaining high performance, which is crucial for efficiency in practical applications.
In summary, MGIR not only demonstrates significant performance improvements but also shows excellent competitiveness when compared with the current state-of-the-art methods. These results highlight the efficiency of MGIR in fixed compression ratio reconstruction, especially in spatial reconstruction tasks.

\noindent\textbf{Qualitative Results.}
\cref{cave_band28} visualizes the reconstructed results for four out of the 28 spectral channels, where MGIR exhibits exceptional visual accuracy. Furthermore, our spectral profile closely aligns with the ground truth, achieving a spectral correlation of 0.997. This represents the highest level of spectral curve correlation compared to other algorithms.

\subsection{Ablation Study}
We utilize the ICVL dataset to perform ablation studies, which are structured to sequentially assess the contributions of key components within our model. Initially, we examine the SSDW module to determine its effectiveness and efficiency. Following this, we proceed with a detailed ablation study to dissect the impact of Positional Encoding (PE) and MGLFA. Finally, we explore the implications of lightweight optimization. To provide a clear demonstration of the performance of each module, these ablation studies are conducted on the version of the model that has not undergone lightweight optimization. This approach allows us to isolate and directly attribute performance gains to the specific enhancements made.

\begin{table*}
\centering
\caption{Comparative performance of 3D Convolution versus our proposed Spectral-Spatial Depthwise Convolution (SSDW).}
\label{tab: different convolution methods}
\setlength{\tabcolsep}{15.0pt}
\begin{tabular}{c|ccc|ccc|ccc}
\toprule
Method    & \multicolumn{3}{c|}{$3\times3\times3$} & \multicolumn{3}{c|}{$5\times5\times5$} & \multicolumn{3}{c}{SSDW} \\
\midrule
Metrics   & RMSE    & PSNR    & SAM   & RMSE    & PSNR    & SAM   & RMSE    & PSNR   & SAM   \\ 
\midrule
31 Bands  &   0.014      &       39.13  &  0.050     &  0.012     & 39.67       &   0.046   & 0.012        &39.84        & 0.045      \\
62 Bands  &  0.019       &    36.48     & 0.095      & 0.018        &  36.71       & 0.088      & 0.016   &      38.12  & 0.070      \\
124 Bands & 0.018        &  36.91      &   0.095 & 0.018        &  37.15       &  0.092    & 0.015        &  38.33      & 0.067      \\
241 Bands &   0.018      & 37.12        &  0.087     &  0.017       &    37.45     &  0.086     & 0.015        &  38.42     & 0.062     \\
\midrule
FLOPs     & \multicolumn{3}{c|}{723.24 $\rm G$}      & \multicolumn{3}{c|}{2801.46 $\rm G$}      & \multicolumn{3}{c}{156.32 $\rm G$}     \\
\# Param   & \multicolumn{3}{c|}{11.03 $\rm M$}      & \multicolumn{3}{c|}{45.15 $\rm M$}      & \multicolumn{3}{c}{1.46 $\rm M$}     \\ 
\bottomrule
\end{tabular}%
\end{table*}

\begin{table*}[t]
\centering
\setlength{\tabcolsep}{9.4pt}
\caption{Ablation study on key components on the ICVL dataset. RPE: Relative Position Embedding, SGLFA: Single-Granularity Local Feature Aggregator, and MGLFA: Mixed-Granularity Local Feature Aggregator. The notation $\times31$-$\times1$ indicates a spectral compression ratio of 31 and a spatial compression ratio of 1.}
\begin{tabular}{cccc|ccc|ccc|ccc}
\toprule
\multirow{2}{*}{Baseline} & \multirow{2}{*}{RPE} & \multirow{2}{*}{SGLFA} & \multirow{2}{*}{MGLFA} & \multicolumn{3}{c|}{$\times$31-$\times$1} & \multicolumn{3}{c|}{$\times$62-$\times$2} & \multicolumn{3}{c}{$\times$241-$\times$4} \\
\cline{5-13}
 &  &  &  & PSNR & SSIM & SAM & PSNR & SSIM & SAM & PSNR & SSIM & SAM \\ 
\midrule
\checkmark &  &  &  & 38.12 & 0.954 & \multicolumn{1}{c|}{0.053} & 34.21 & 0.927 & \multicolumn{1}{c|}{0.091} & 31.75 & 0.902 & 0.094 \\
\checkmark & \checkmark  &  &  & 38.20 & 0.953 & \multicolumn{1}{c|}{0.056} & 34.46 & 0.929 & \multicolumn{1}{c|}{0.090} & 31.82 & 0.903 & 0.089 \\
\checkmark  & \checkmark  & \checkmark  &  & 38.82 & 0.959 & \multicolumn{1}{c|}{0.050} & 34.52 & 0.931 & \multicolumn{1}{c|}{0.087} & 32.19 & 0.907 & 0.101 \\
\checkmark  & \checkmark  & \checkmark  & \checkmark  & 39.84 & 0.965 & \multicolumn{1}{c|}{0.045} & 35.52 & 0.941 & \multicolumn{1}{c|}{0.084} & 32.69 & 0.914 & 0.088 \\
\bottomrule
\end{tabular}
\label{ablation study}
\end{table*}

\begin{figure}
\centering
\includegraphics[width=\linewidth]{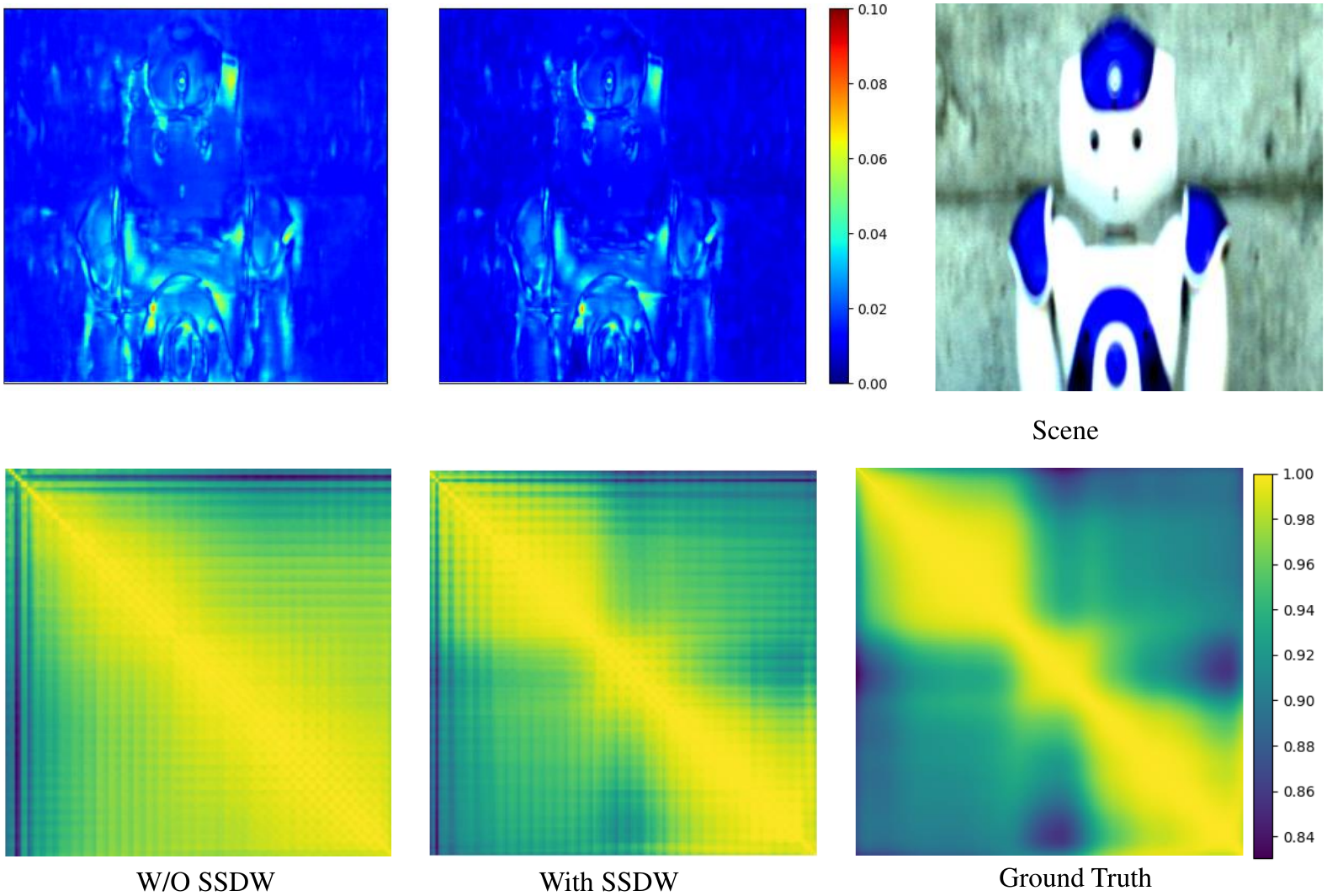}
\caption{Visual comparison illustrating the effects of implementing versus omitting Spectral-Spatial Depthwise Convolution.}
\label{ab_ssdw}
\end{figure}

\begin{figure}[t]
\centering
\includegraphics[width=0.85\linewidth]{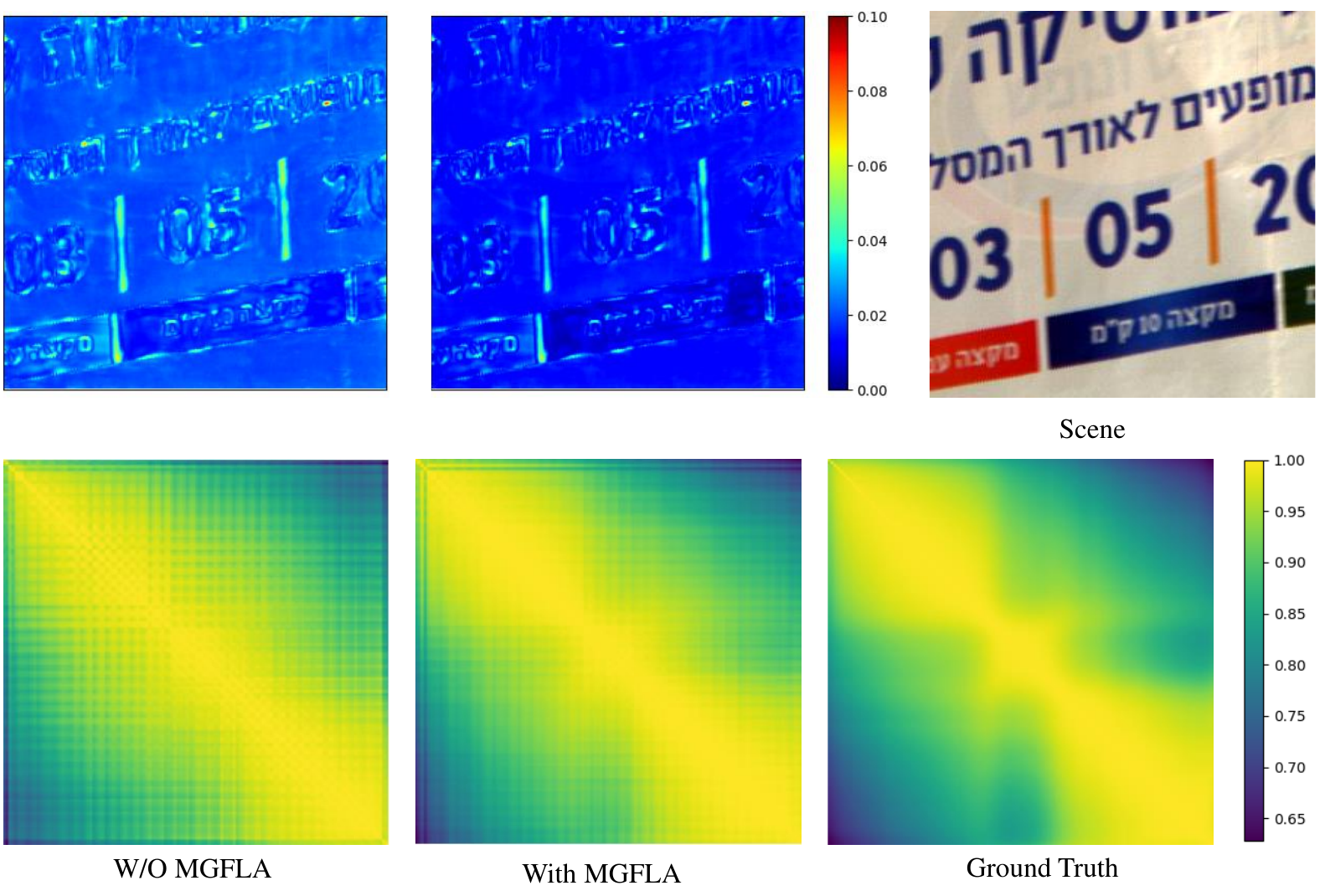}
\caption{Visual comparison demonstrating the presence versus absence of the Mixed-Granularity Local Feature Aggregator.}
\label{ab_mgfla}
\end{figure}

\begin{table*}[t]
\centering
\caption{Ablation study on the number of attention heads in the Mixed-Granularity Local Feature Aggregator (MGFLA).}
\label{tab: number of multi-heads}
\setlength{\tabcolsep}{9.0pt}
\begin{tabular}{c|ccc|ccc|ccc|ccc}
\toprule
Number    & \multicolumn{3}{c|}{1 Head} & \multicolumn{3}{c|}{2 Heads} & \multicolumn{3}{c|}{3 Heads} & \multicolumn{3}{c}{4 Heads} \\
\midrule
Metrics   & RMSE   & PSNR  & SAM  & RMSE   & PSNR  & SAM  & RMSE   & PSNR  & SAM  & RMSE   & PSNR  & SAM  \\ 
\midrule
31 Bands  &  0.014   &  38.82    & 0.050     & 0.013       &     39.41  &  0.047    & 0.013       &  39.66    &0.046      &0.012        & 39.84      &0.045      \\
62 Bands  & 0.017       &  37.51     &   0.073   &  0.016      &   37.88    &  0.071&  0.016     &   38.01&0.071   &  0.016      &   38.12    &   0.070   \\
124 Bands & 0.016 &  37.63  &  0.073    & 0.016       & 38.03      & 0.070     &  0.015    &  38.12     &  0.068    & 0.015       &38.33       &  0.067    \\
241 Bands & 0.016       &  37.77     & 0.067     & 0.015       &   38.14    &    0.063  &   0.015     &   38.22    &  0.063    &   0.015     &   38.42    &  0.062    \\ 
\bottomrule
\end{tabular}%
\end{table*}

\begin{table}[t]
\centering
\caption{Experimental results for various feature fusion operations.}
\label{tab: feature fusion operation}
\renewcommand\arraystretch{1.1}
\setlength{\tabcolsep}{6.0pt}
\begin{tabular}{c|ccc|ccc}
\toprule
Operation & \multicolumn{3}{c|}{Concatenation} & \multicolumn{3}{c}{Addition} \\
\midrule
Metrics   & RMSE       & PSNR      & SAM      & RMSE     & PSNR     & SAM    \\ \midrule
31 Bands  & 0.015          &  38.32        &0.065          &0.012          & 39.84         & 0.045       \\
62 Bands  &    0.018       &   37.17        &  0.082        &  0.016        & 38.12         & 0.070       \\
124 Bands &     0.017  & 37.54  &  0.087        & 0.015         &  38.33        & 0.067       \\
241 Bands &      0.017     &   37.51        &    0.083      & 0.015         &  38.42        &   0.062     \\ 
\bottomrule
\end{tabular}%
\end{table}

\begin{figure}[t]
\centering
\includegraphics[width=\linewidth]{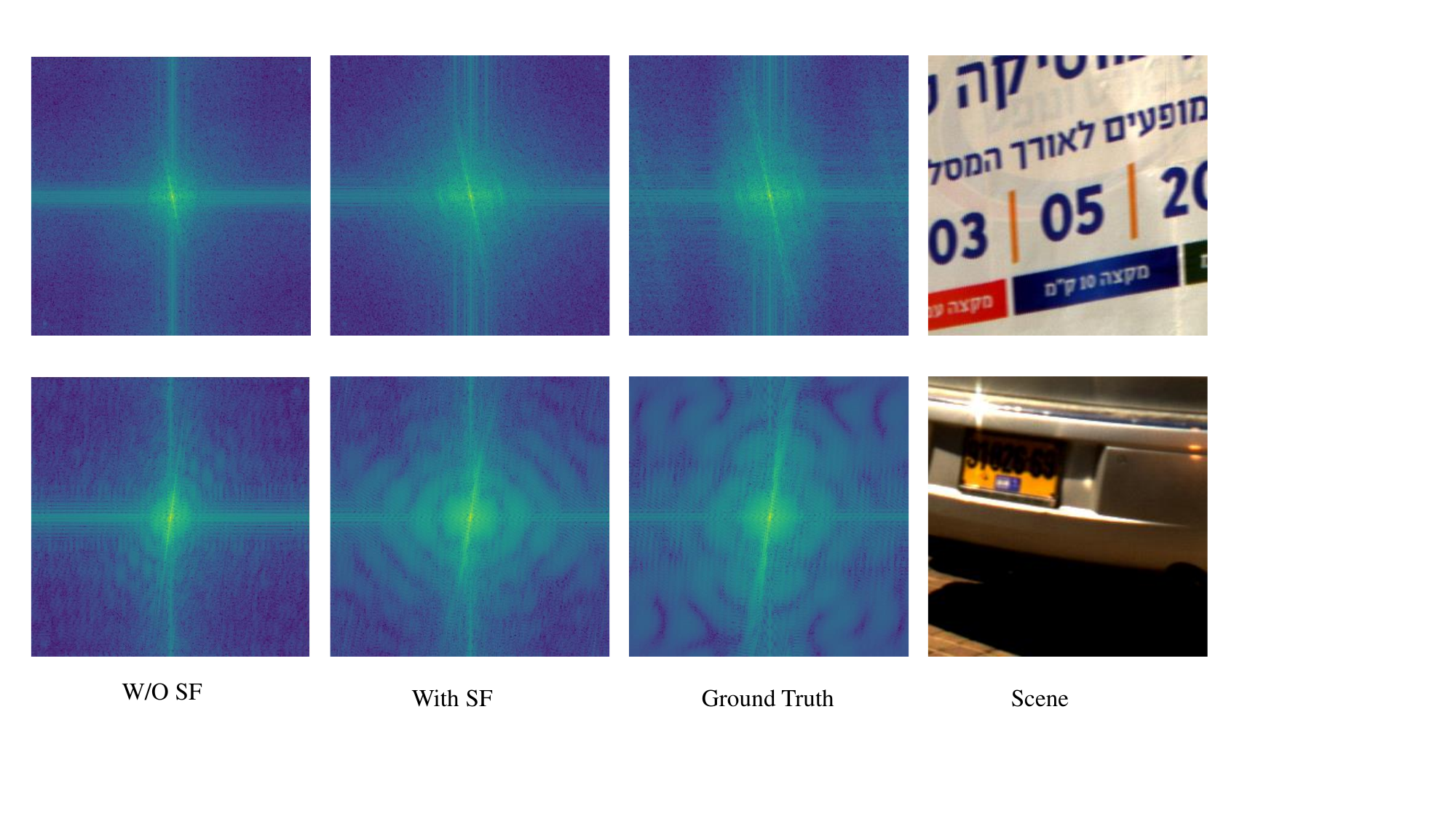}
\caption{Visual comparison illustrating the presence and absence of 3D coordinate mapping with Sinusoidal Functions (SF).}
\label{ab_sf}
\end{figure}



\noindent\textbf{Effect of Spectral-Spatial Depthwise Convolution.}
We conducted an ablation study on the Spectral-Spatial Depthwise Convolution (SSDW) module to assess its efficacy by comparing it with Global Multi-head Self-Attention (G-MSA) and Window-based Multi-head Self-Attention (W-MSA). The results, detailed in~\cref{tab:light}, demonstrate that SSDW consistently outperformed both G-MSA and W-MSA across nearly all tested compression ratios. Moreover, SSDW achieved these superior performance with fewer FLOPs and a reduced parameter count. These findings underscore the efficiency of our SSDW module in capturing spectral correlations with minimal computational overhead.

We further evaluate the SSDW module by comparing it with conventional $3\times3\times3$ and $5\times5\times5$ convolutions. As shown in~\cref{tab: different convolution methods}, SSDW consistently outperforms these traditional methods across all evaluated spectral band reconstruction scenarios. In terms of computational complexity, SSDW achieves a significant reduction in FLOPs, with 77.79\% and 94.42\% fewer operations compared to $3\times3\times3$ and $5\times5\times5$ convolutions, respectively. Moreover, SSDW maintains a minimal parameter count of 1.46M. These results highlight the superior efficiency and effectiveness of the SSDW module.

\cref{ab_ssdw} displays reconstruction error maps and spectral correlation coefficient maps, comparing scenarios with and without the SSDW module. The error maps illustrate the average error across all 124 bands, revealing that the images reconstructed with SSDW exhibit consistently lower errors. This improvement is evident regardless of the size of structural details or the boundaries between different objects. Additionally, the spectral correlation coefficient maps show that the results using SSDW more closely align with the ground truth across the spectral bands, further demonstrating the effectiveness of the SSDW module.

\noindent {\bf{Effect of Mixed-granularity Feature Aggregator.}} 
To assess the effectiveness of the Mixed-Granularity Local Feature Aggregator, we compared it with a simpler model variant, SGLFA, which accesses only the single-scale, uppermost features in the Hierarchical Spectral-Spatial Implicit Encoder. As detailed in~\cref{ablation study}, incorporating SGLFA into the model yields a 0.62 $\rm dB$ improvement at a spectral compression ratio of 1. Introducing MGLFA, however, raises the PSNR from 38.82 $\rm dB$ to 39.84 $\rm dB$ at single granularity, achieving a 1.02 $\rm dB$ increase. Furthermore, when employing multiple granularities in the feature aggregator, the PSNR enhancement ranges from 0.94 to 1.72 $\rm dB$ above the baseline. These results confirm that adaptively aggregating diverse fine-grained features significantly improves the Image Neural Representation, with noticeable gains across various compression scales.

\cref{ab_mgfla} further evaluates the reconstruction results with and without the Mixed-Granularity Feature Aggregator. The error maps distinctly show that integrating the aggregator significantly reduces overall reconstruction error and notably enhances detail representation.

We also conduct an ablation study on the number of heads in the Multi-granularity Local Feature Aggregator.~\cref{tab: number of multi-heads} shows that varying the number of heads affects the reconstruction results. Increasing the number of heads from 1 to 4 in the case of 31 spectral bands boosts the PSNR from 38.82 to 39.84 $\rm dB$ and reduces the SAM from 0.050 to 0.045. Similar improvements were observed with other numbers of spectral bands, confirming that more heads effectively enhance reconstruction accuracy by enriching feature granularity.

Additionally, we explore feature fusion methods, including Concatenation and Addition, to integrate features of varying scales. As indicated in~\cref{tab: feature fusion operation}, element-wise Addition consistently surpasses Concatenation in reconstructing different spectral bands.

\begin{table}[t]
\centering
\caption{Comparative analysis of 3D coordinate mapping using sinusoidal functions versus without.}
\label{tab: 3D-C}
\renewcommand\arraystretch{1.1}
\setlength{\tabcolsep}{6.0pt}
\begin{tabular}{c|ccc|ccc}
\toprule
Operation & \multicolumn{3}{c|}{w/o Sinusoidal Function} & \multicolumn{3}{c}{with Sinusoidal Function} \\
\midrule
Metrics   & RMSE       & PSNR      & SAM      & RMSE     & PSNR     & SAM    \\ \midrule
31 Bands  & 0.015         &  38.29        &0.057          &0.012          & 39.84         & 0.045       \\
62 Bands  &    0.017       &   37.46        &  0.080        &  0.016        & 38.12         & 0.070       \\
124 Bands &     0.017  & 37.50  &  0.076        & 0.015         &  38.33        & 0.067       \\
241 Bands &      0.017     &   37.43        &    0.073      & 0.015         &  38.42        &   0.062     \\ 
\bottomrule
\end{tabular}%
\vspace{-0.4cm}
\end{table}

\noindent {\bf{Effect of Relative Positional Encoding.}} 
In~\cref{ablation study}, the term $\text{baseline}$ refers to a method that amplifies the number of bands obtained by interpolating post-reconstruction from the backbone. PE denotes the addition of relative position information to the baseline. According to the table, the baseline model at a $\times 31$ compression yields a PSNR of 38.12 $\rm dB$, and the accuracy of both spatial and spectral reconstructions improves as the number of reconstructed spectra increases. Further, we conducted additional ablation studies to assess the contribution of the Sinusoidal Function, with results now presented in~\cref{tab: 3D-C}. The integration of relative position encoding has demonstrated a significant improvement, with a maximum increase of 1.55 $\rm dB$ in PSNR and 0.12 in SAM. These enhancements reveal that relative position encoding effectively orchestrates feature aggregation across various granularities, leading to superior reconstruction .

Moreover,~\cref{ab_sf} details an average frequency domain visualization for two samples across 124 bands. Incorporating sinusoidal functions for positional encoding allowed the frequencies of the reconstructed images to align more closely with those of the original images, particularly excelling in the reconstruction of high-frequency details. This confirms that mapping coordinates to higher dimensions using sinusoidal functions significantly enhances the reconstruction quality of high-frequency details.

\begin{table*}[t]
\centering
\caption{Quantitative results for continuous spectral reconstruction on the ICVL dataset. MGIR-A refers to the model prior to lightweight optimization, while MGIR denotes our latest version. }
\renewcommand\arraystretch{1.1}
\setlength{\tabcolsep}{9.0pt}
\begin{tabular}{l|ccc|ccc|ccc|ccc}
\toprule
\multirow{2}{*}{Method} & \multicolumn{3}{c|}{31 Bands} & \multicolumn{3}{c|}{62 Bands} & \multicolumn{3}{c|}{124 Bands} & \multicolumn{3}{c}{241 Bands} \\ \cline{2-13}
 & \multicolumn{1}{c}{RMSE} & \multicolumn{1}{c}{PSNR} & \multicolumn{1}{c|}{SAM} & \multicolumn{1}{c}{RMSE} & \multicolumn{1}{c}{PSNR} & \multicolumn{1}{c|}{ SAM} & \multicolumn{1}{c}{RMSE } & \multicolumn{1}{c}{PSNR} & \multicolumn{1}{c|}{SAM} & \multicolumn{1}{c}{RMSE} & \multicolumn{1}{c}{PSNR} & \multicolumn{1}{c}{SAM} \\ 
 \midrule
MGIR-A & 0.012 & 39.84 & 0.045 & 0.016 & 38.12 & 0.070 & 0.015 & 38.33 & 0.067 & 0.015 & 38.42 &  0.062\\ 
MGIR &\textbf{0.011}  & \textbf{40.21} &\textbf{0.041}  &\textbf{0.015}  &\textbf{38.51} &\textbf{0.065} &\textbf{0.015} &\textbf{38.67} &\textbf{0.058} &\textbf{0.014} &\textbf{39.07} &\textbf{0.054}   \\ 
\bottomrule
\end{tabular}
\label{tab:compare_ICVL}
\end{table*}

\begin{table*}
\centering
\caption{Quantitative results of spatial reconstruction of 10 scenes on KAIST. PSNR/SSIM (the first/second line) are reported.}
\label{compare_cave_data}
\renewcommand\arraystretch{1.1}
\setlength{\tabcolsep}{8pt}
\begin{tabular}{ l | c  c | c  c  c  c  c  c  c  c  c  c  | c c}
\toprule
Methods & \#Params & FLOPs & S1 & S2 & S3 & S4 & S5 & S6 & S7 & S8 & S9 & S10 & Average \\
\midrule
\multirow{2}{*}{MGIR-A}
&\multirow{2}{*}{1.45  $\rm M$} & \multirow{2}{*}{144.4 $\rm G$}& 34.78  & 34.32& 37.71 & 43.26 & 33.38 & 33.23 & 34.36 & 31.70 & \bf{36.48}& 30.51&34.97\\
& & & 0.924 & 0.911 & \bf{0.961} & 0.963 & 0.934 & 0.931 & 0.944 & 0.923 & 0.941 & 0.883 & 0.932\\

\multirow{2}{*}{MGIR}
&\multirow{2}{*}{0.92  $\rm M$} & \multirow{2}{*}{33.05 $\rm G$} &\textbf{35.72}  &\textbf{35.29}  &\textbf{37.82}  &\textbf{43.43}  &\textbf{33.75}  &\textbf{34.15}  &\textbf{34.52}  &\textbf{33.64}  &36.26  &\textbf{32.34}  &\textbf{35.69} \\  
& & &\textbf{0.941}   &\textbf{0.943}   &0.956  &\textbf{0.980}  &\textbf{0.952}  &\textbf{0.948}  &\textbf{0.955}  &\textbf{0.947}  &\textbf{0.951}  & \textbf{0.942}  & \textbf{0.951}
\\ 

\bottomrule
\end{tabular}
\end{table*}

\noindent {\bf{Effect of Lightweight Optimization.}} 
To enhance model efficiency, we implement lightweight optimization on our model. We replace the 3D convolutional layers with more efficient 3D depth-wise separable convolutions. Additionally, we enhance the feature extraction capability of the encoder by increasing the number of stacked SSDW modules from (2,2,2,2) to (2,2,4,4), which has a minimal impact on computational resources. Moreover, to improve the model's expressive power, we adjust the network feature dimensions, setting the encoder's stage feature channels to $C_1=32$ and $C_2=16$, and increase the feature dimension of the MGLFA module from 32 to 64 to better adapt to multi-scale features. This results in a 36.5\% reduction in the number of parameters (from 1.45  $\rm M$ to 0.92  $\rm M$) and a 77.0\% decrease in FLOPs (from 144.4 $\rm G$ to 33.05  $\rm G$), maintaining performance compared to the unoptimized MGIR-A, thus realizing a more efficient model structure.

Experimental results on the KAIST dataset further confirm the significant advantages of the lightweight optimized MGIR over the MGIR-A. As shown in~\cref{compare_cave_data}, MGIR improve the average PSNR and SSIM metrics by 0.72 $\rm dB$ and 0.019, respectively, indicating enhanced generalization capabilities and reconstruction accuracy across different scenes. Specifically, the average PSNR of MGIR reached 35.69 $\rm dB$, and the SSIM is 0.951, which is 0.51 $\rm dB$ higher in PSNR and 0.003 better in SAM compared to the best-performing MST-L model in the MST series. These results demonstrate that the lightweight optimized MGIR surpasses the non-optimized version in both computational complexity and performance, showing a significant efficiency advantage.

The results further demonstrate the significant advantages of our MGIR over the unoptimized version MGIR-A. As shown in~\cref{tab:compare_ICVL}, on the ICVL dataset, MGIR outperforms MGIR-A in continuous spectral reconstruction tasks across all bands, particularly in terms of PSNR and SAM metrics. Especially at the 241st band, MGIR's PSNR is 0.65 $\rm dB$ higher and SAM is improved by 0.008 compared to MGIR-A, highlighting the advantage of our continuous reconstruction strategy in handling high-band images.

\section{Conclusion}
This study introduces a mixed-granularity implicit representation for continuous hyperspectral compressive reconstruction, enabling reconstructed hyperspectral images to be displayed at any controllable spectral spatial resolution. This is the first comprehensive effort to integrate implicit neural representation with CASSI reconstruction. We conceptualize hyperspectral images as continuous functions of spectral-spatial coordinates and develop a hierarchical encoder to extract multi-scale latent codes. Additionally, we design a local feature aggregator with a multi-headed attention mechanism to cluster features across scales efficiently. Both qualitative and quantitative assessments show that our model produces visually appealing and high-quality reconstructed hyperspectral images.

\bibliographystyle{IEEEtran}
\bibliography{main}

\begin{thebibliography}{10}
\providecommand{\url}[1]{#1}
\csname url@samestyle\endcsname
\providecommand{\newblock}{\relax}
\providecommand{\bibinfo}[2]{#2}
\providecommand{\BIBentrySTDinterwordspacing}{\spaceskip=0pt\relax}
\providecommand{\BIBentryALTinterwordstretchfactor}{4}
\providecommand{\BIBentryALTinterwordspacing}{\spaceskip=\fontdimen2\font plus
\BIBentryALTinterwordstretchfactor\fontdimen3\font minus \fontdimen4\font\relax}
\providecommand{\BIBforeignlanguage}[2]{{%
\expandafter\ifx\csname l@#1\endcsname\relax
\typeout{** WARNING: IEEEtran.bst: No hyphenation pattern has been}%
\typeout{** loaded for the language `#1'. Using the pattern for}%
\typeout{** the default language instead.}%
\else
\language=\csname l@#1\endcsname
\fi
#2}}
\providecommand{\BIBdecl}{\relax}
\BIBdecl

\bibitem{jolly2022hyperspectral}
S.~W. Jolly, ``Hyperspectral imaging and pulse characterization,'' \emph{Light: Science \& Applications}, vol.~11, no.~1, p. 267, 2022.

\bibitem{10313066}
P.~Liu, T.~Xu, H.~Chen, S.~Zhou, H.~Qin, and J.~Li, ``Spectrum-driven mixed-frequency network for hyperspectral salient object detection,'' \emph{IEEE Transactions on Multimedia}, vol.~26, pp. 5296--5310, 2024.

\bibitem{10475351}
H.~Qin, T.~Xu, P.~Liu, J.~Xu, and J.~Li, ``Dmssn: Distilled mixed spectral–spatial network for hyperspectral salient object detection,'' \emph{IEEE Transactions on Geoscience and Remote Sensing}, vol.~62, pp. 1--18, 2024.

\bibitem{zhang2021deeply}
W.~Zhang, H.~Song, X.~He, L.~Huang, X.~Zhang, J.~Zheng, W.~Shen, X.~Hao, and X.~Liu, ``Deeply learned broadband encoding stochastic hyperspectral imaging,'' \emph{Light: Science \& Applications}, vol.~10, no.~1, pp. 1--7, 2021.

\bibitem{tang2022single}
H.~Tang, T.~Men, X.~Liu, Y.~Hu, J.~Su, Y.~Zuo, P.~Li, J.~Liang, M.~C. Downer, and Z.~Li, ``Single-shot compressed optical field topography,'' \emph{Light: Science \& Applications}, vol.~11, no.~1, pp. 1--10, 2022.

\bibitem{10261266}
H.~Chen, W.~Zhao, T.~Xu, G.~Shi, S.~Zhou, P.~Liu, and J.~Li, ``Spectral-wise implicit neural representation for hyperspectral image reconstruction,'' \emph{IEEE Transactions on Circuits and Systems for Video Technology}, vol.~34, no.~5, pp. 3714--3727, 2024.

\bibitem{cai2022mask}
Y.~Cai, J.~Lin, X.~Hu, H.~Wang, X.~Yuan, Y.~Zhang, R.~Timofte, and L.~Van~Gool, ``Mask-guided spectral-wise transformer for efficient hyperspectral image reconstruction,'' in \emph{Proceedings of the IEEE/CVF Conference on Computer Vision and Pattern Recognition}, 2022, pp. 17\,502--17\,511.

\bibitem{wang2021tsa}
S.~Wang, D.~Yang, P.~Zhai, C.~Chen, and L.~Zhang, ``Tsa-net: Tube self-attention network for action quality assessment,'' in \emph{Proceedings of the 29th ACM International Conference on Multimedia}, 2021, pp. 4902--4910.

\bibitem{meng2020gap}
Z.~Meng, S.~Jalali, and X.~Yuan, ``Gap-net for snapshot compressive imaging,'' \emph{arXiv preprint arXiv:2012.08364}, 2020.

\bibitem{cai2022degradation}
Y.~Cai, J.~Lin, H.~Wang, X.~Yuan, H.~Ding, Y.~Zhang, R.~Timofte, and L.~V. Gool, ``Degradation-aware unfolding half-shuffle transformer for spectral compressive imaging,'' \emph{Advances in Neural Information Processing Systems}, vol.~35, pp. 37\,749--37\,761, 2022.

\bibitem{zhang2021holistic}
C.~Zhang, Z.~Cui, Y.~Zhang, B.~Zeng, M.~Pollefeys, and S.~Liu, ``Holistic 3d scene understanding from a single image with implicit representation,'' in \emph{Proceedings of the IEEE/CVF Conference on Computer Vision and Pattern Recognition}, 2021, pp. 8833--8842.

\bibitem{du2021neural}
Y.~Du, Y.~Zhang, H.-X. Yu, J.~B. Tenenbaum, and J.~Wu, ``Neural radiance flow for 4d view synthesis and video processing,'' in \emph{2021 IEEE/CVF International Conference on Computer Vision}, 2021, pp. 14\,304--14\,314.

\bibitem{wang2021neus}
P.~Wang, L.~Liu, Y.~Liu, C.~Theobalt, T.~Komura, and W.~Wang, ``Neus: Learning neural implicit surfaces by volume rendering for multi-view reconstruction,'' \emph{arXiv preprint arXiv:2106.10689}, 2021.

\bibitem{chen2021learning}
Y.~Chen, S.~Liu, and X.~Wang, ``Learning continuous image representation with local implicit image function,'' in \emph{Proceedings of the IEEE/CVF conference on computer vision and pattern recognition}, 2021, pp. 8628--8638.

\bibitem{miao2019net}
X.~Miao, X.~Yuan, Y.~Pu, and V.~Athitsos, ``l-net: Reconstruct hyperspectral images from a snapshot measurement,'' in \emph{Proceedings of the IEEE/CVF International Conference on Computer Vision}, 2019, pp. 4059--4069.

\bibitem{10017363}
S.~Zhou, T.~Xu, S.~Dong, and J.~Li, ``Rdfnet: Regional dynamic fista-net for spectral snapshot compressive imaging,'' \emph{IEEE Transactions on Computational Imaging}, vol.~9, pp. 490--501, 2023.

\bibitem{sitzmann2019scene}
V.~Sitzmann, M.~Zollh{\"o}fer, and G.~Wetzstein, ``Scene representation networks: Continuous 3d-structure-aware neural scene representations,'' \emph{Advances in Neural Information Processing Systems}, vol.~32, 2019.

\bibitem{jiang2020local}
C.~Jiang, A.~Sud, A.~Makadia, J.~Huang, M.~Nie{\ss}ner, T.~Funkhouser \emph{et~al.}, ``Local implicit grid representations for 3d scenes,'' in \emph{Proceedings of the IEEE/CVF Conference on Computer Vision and Pattern Recognition}, 2020, pp. 6001--6010.

\bibitem{mildenhall2021nerf}
B.~Mildenhall, P.~P. Srinivasan, M.~Tancik, J.~T. Barron, R.~Ramamoorthi, and R.~Ng, ``Nerf: Representing scenes as neural radiance fields for view synthesis,'' \emph{Communications of the ACM}, vol.~65, no.~1, pp. 99--106, 2021.

\bibitem{han2021transformer}
K.~Han, A.~Xiao, E.~Wu, J.~Guo, C.~Xu, and Y.~Wang, ``Transformer in transformer,'' \emph{Advances in Neural Information Processing Systems}, vol.~34, pp. 15\,908--15\,919, 2021.

\bibitem{wang2022uformer}
Z.~Wang, X.~Cun, J.~Bao, W.~Zhou, J.~Liu, and H.~Li, ``Uformer: A general u-shaped transformer for image restoration,'' in \emph{Proceedings of the IEEE/CVF Conference on Computer Vision and Pattern Recognition}, 2022, pp. 17\,683--17\,693.

\bibitem{liu2021swin}
Z.~Liu, Y.~Lin, Y.~Cao, H.~Hu, Y.~Wei, Z.~Zhang, S.~Lin, and B.~Guo, ``Swin transformer: Hierarchical vision transformer using shifted windows,'' in \emph{Proceedings of the IEEE/CVF International Conference on Computer Vision}, 2021, pp. 10\,012--10\,022.

\bibitem{ren2021shunted}
S.~Ren, D.~Zhou, S.~He, J.~Feng, and X.~Wang, ``Shunted self-attention via multi-scale token aggregation,'' 2021.

\bibitem{chen2021crossvit}
C.-F.~R. Chen, Q.~Fan, and R.~Panda, ``Crossvit: Cross-attention multi-scale vision transformer for image classification,'' in \emph{Proceedings of the IEEE/CVF international conference on computer vision}, 2021, pp. 357--366.

\bibitem{lee2022local}
J.~Lee and K.~H. Jin, ``Local texture estimator for implicit representation function,'' in \emph{Proceedings of the IEEE/CVF Conference on Computer Vision and Pattern Recognition}, 2022, pp. 1929--1938.

\bibitem{dosovitskiy2020image}
A.~Dosovitskiy, L.~Beyer, A.~Kolesnikov, D.~Weissenborn, X.~Zhai, T.~Unterthiner, M.~Dehghani, M.~Minderer, G.~Heigold, S.~Gelly \emph{et~al.}, ``An image is worth 16x16 words: Transformers for image recognition at scale,'' \emph{arXiv preprint arXiv:2010.11929}, 2020.

\bibitem{vaswani2017attention}
A.~Vaswani, N.~Shazeer, N.~Parmar, J.~Uszkoreit, L.~Jones, A.~N. Gomez, {\L}.~Kaiser, and I.~Polosukhin, ``Attention is all you need,'' \emph{Advances in neural information processing systems}, vol.~30, 2017.

\bibitem{voita2019analyzing}
E.~Voita, D.~Talbot, F.~Moiseev, R.~Sennrich, and I.~Titov, ``Analyzing multi-head self-attention: Specialized heads do the heavy lifting, the rest can be pruned,'' \emph{arXiv preprint arXiv:1905.09418}, 2019.

\bibitem{esmaeilzadeh2020meshfreeflownet}
S.~Esmaeilzadeh, K.~Azizzadenesheli, K.~Kashinath, M.~Mustafa, H.~A. Tchelepi, P.~Marcus, M.~Prabhat, A.~Anandkumar \emph{et~al.}, ``Meshfreeflownet: A physics-constrained deep continuous space-time super-resolution framework,'' in \emph{SC20: International Conference for High Performance Computing, Networking, Storage and Analysis}.\hskip 1em plus 0.5em minus 0.4em\relax IEEE, 2020, pp. 1--15.

\bibitem{cst}
Y.~Cai, J.~Lin, X.~Hu, H.~Wang, X.~Yuan, Y.~Zhang, R.~Timofte, and L.~V. Gool, ``Coarse-to-fine sparse transformer for hyperspectral image reconstruction,'' in \emph{European Conference on Computer Vision}, 2022.

\bibitem{kingma2014adam}
D.~P. Kingma and J.~Ba, ``Adam: A method for stochastic optimization,'' \emph{arXiv preprint arXiv:1412.6980}, 2014.

\bibitem{arad_and_ben_shahar_2016_ECCV}
B.~Arad and O.~Ben-Shahar, ``Sparse recovery of hyperspectral signal from natural rgb images,'' in \emph{European Conference on Computer Vision}.\hskip 1em plus 0.5em minus 0.4em\relax Springer, 2016, pp. 19--34.

\bibitem{Huang_2017}
\BIBentryALTinterwordspacing
J.~Huang and D.~Shi, ``Multispectral computational ghost imaging with multiplexed illumination,'' \emph{Journal of Optics}, vol.~19, no.~7, p. 075701, jun 2017. [Online]. Available: \url{https://doi.org/10.10882F2040-89862Faa72ff}
\BIBentrySTDinterwordspacing

\bibitem{choi2017high}
I.~Choi, M.~Kim, D.~Gutierrez, D.~Jeon, and G.~Nam, ``High-quality hyperspectral reconstruction using a spectral prior,'' Tech. Rep., 2017.

\bibitem{hu2022hdnet}
X.~Hu, Y.~Cai, J.~Lin, H.~Wang, X.~Yuan, Y.~Zhang, R.~Timofte, and L.~Van~Gool, ``Hdnet: High-resolution dual-domain learning for spectral compressive imaging,'' in \emph{Proceedings of the IEEE/CVF Conference on Computer Vision and Pattern Recognition}, 2022, pp. 17\,542--17\,551.

\bibitem{DBLP:journals/corr/Larkin15}
\BIBentryALTinterwordspacing
K.~G. Larkin, ``Structural similarity index ssimplified: Is there really a simpler concept at the heart of image quality measurement?'' \emph{CoRR}, vol. abs/1503.06680, 2015. [Online]. Available: \url{http://arxiv.org/abs/1503.06680}
\BIBentrySTDinterwordspacing

\bibitem{kruse1993spectral}
F.~A. Kruse, A.~Lefkoff, J.~Boardman, K.~Heidebrecht, A.~Shapiro, P.~Barloon, and A.~Goetz, ``The spectral image processing system (sips)—interactive visualization and analysis of imaging spectrometer data,'' \emph{Remote sensing of environment}, vol.~44, no. 2-3, pp. 145--163, 1993.

\bibitem{meng2020end}
Z.~Meng, J.~Ma, and X.~Yuan, ``End-to-end low cost compressive spectral imaging with spatial-spectral self-attention,'' in \emph{European Conference on Computer Vision}.\hskip 1em plus 0.5em minus 0.4em\relax Springer, 2020, pp. 187--204.

\bibitem{meng2023deep}
Z.~Meng, X.~Yuan, and S.~Jalali, ``Deep unfolding for snapshot compressive imaging,'' \emph{International Journal of Computer Vision}, vol. 131, no.~11, pp. 2933--2958, 2023.

\bibitem{9879297}
X.~Hu, Y.~Cai, J.~Lin, H.~Wang, X.~Yuan, Y.~Zhang, R.~Timofte, and L.~Van~Gool, ``Hdnet: High-resolution dual-domain learning for spectral compressive imaging,'' in \emph{CVPR}, 2022.

\bibitem{mst}
Y.~Cai, J.~Lin, X.~Hu, H.~Wang, X.~Yuan, Y.~Zhang, R.~Timofte, and L.~V. Gool, ``Mask-guided spectral-wise transformer for efficient hyperspectral image reconstruction,'' in \emph{Proceedings of the IEEE/CVF conference on computer vision and pattern recognition}, 2022.

\end{thebibliography}

\end{document}